\documentclass{article}

\usepackage{PRIMEarxiv}
\usepackage[utf8]{inputenc} %
\usepackage[T1]{fontenc}    %
\usepackage{hyperref}       %
\usepackage{url}            %
\usepackage{booktabs}       %
\usepackage{amsfonts}       %
\usepackage{nicefrac}       %
\usepackage{microtype}      %
\usepackage{lipsum}
\usepackage{fancyhdr}       %
\usepackage{graphicx}       %

\usepackage{xcolor}
\definecolor{newcolor}{rgb}{.8,.349,.1}

\usepackage{tabularx}
\usepackage{standalone}
\usepackage{gensymb}
\usepackage{mathpazo}
\usepackage{longtable}
\usepackage[normalem]{ulem}
\usepackage{float}
\usepackage{svg}
\usepackage{breakurl}
\usepackage{refcount}
\usepackage[flushleft]{threeparttable}
\DeclareMathAlphabet{\pazocal}{OMS}{zplm}{m}{n}

\newcommand{\Lb}{\pazocal{L}}
\usepackage[utf8]{inputenc} %
\usepackage{amsmath,amsthm,amsfonts,dsfont,mathtools} %
\usepackage{longtable}
\def\BibTeX{{\rm B\kern-.05em{\sc i\kern-.025em b}\kern-.08emT\kern-.1667em\lower.7ex\hbox{E}\kern-.125emX}}
\definecolor{lightgray}{gray}{0.9}

\title{ISP meets Deep Learning: A Survey on Deep Learning Methods for Image Signal Processing}

\author{
	Matheus Henrique Marques da Silva, Jhessica Victoria Santos da Silva, Rodrigo Reis Arrais \\
	Eldorado Research Institute \\
	Campinas, São Paulo - Brazil\\
	\texttt{\{matheus.marques, jhessica.silva, rodrigo.arrais\}@eldorado.org.br}
	\And
	Wladimir Barroso Guedes de Ara\'{u}jo Neto, Leonardo Tadeu Lopes, Guilherme Augusto Bileki \\
	Eldorado Research Institute \\
	Campinas, São Paulo - Brazil\\
	\texttt{\{wladimir.neto, leonardo.lopes, bilekig,\}@eldorado.org.br}
	\And
	Iago Oliveira Lima, Lucas Borges Rondon, Bruno Melo de Souza\\
	Eldorado Research Institute \\
	Campinas, São Paulo - Brazil\\
	\texttt{\{iagolima, lucas.rondon, brunobms\}@eldorado.org.br}
	\And
	Mayara Costa Regazio, Rodolfo Coelho Dalapicola, Claudio Filipi Gon\c{c}alves dos Santos\\
	Eldorado Research Institute \\
	Campinas, São Paulo - Brazil\\
	\texttt{\{mayara.regazio, rodolfo.dalapicola, claudio.santos\}@eldorado.org.br}
}

\newcommand{\rodrigo}[1]{\leavevmode\color{black}{#1}}
\newcommand{\jhessica}[1]{\leavevmode\color{black}{#1}}
\newcommand{\matheus}[1]{\leavevmode\color{black}{#1}}

\newcommand{\guilhermeHepfener}[1]{\leavevmode\color{black}{#1}}

\newcommand{\wladimir}[1]{\leavevmode\color{black}{#1}}

\newcommand{\brunoiago}[1]{\leavevmode\color{black}{#1}}

\begin{document}
\maketitle

\begin{abstract}
The entire Image Signal Processor (ISP) of a camera relies on several processes to transform the data from the Color Filter Array (CFA) sensor, such as demosaicing, denoising, and enhancement. These processes can be executed either by some hardware or via software. In recent years, Deep Learning has emerged as one solution for some of them or even to replace the entire ISP using a single neural network for the task. In this work, we investigated several recent pieces of research in this area and provide deeper analysis and comparison among them, including results and possible points of improvement for future researchers.
\end{abstract}


\newpage
\section{Introduction}
\label{s.introduction}

The Image Signal Processor (ISP) is a component of digital cameras capable of performing
various tasks to improve image quality, as demosaicing, denoising, and white balance.
The set of tasks performed by the ISP is called ISP pipeline, divided in preproccessing and postprocessing
steps, and may differ from manufacturer to manufacturer~\cite{ramanath2005Processing}.
Nowadays, Machine Learning is used to replace partially or the entire ISP pipeline. Particulary, Deep Learning is employed
to replace ISP tasks, working on noise removal or some image feaure that hinders processing over the network.
Deep Learning network provides an improvement in relation to computational efficiency and processing time.
This survey paper aims to analyze recent studies, 27 research papers, that implemented
Deep Learning based ISP pipeline.

\subsection{Image Signal Processing}
\label{ss.isp}

Traditionally, ISPs are digital signal processors that reconstruct RGB images from RAW images. In traditional camera pipelines, complex and proprietary hardware processes are used to perform image signal processing ~\cite{zhuyu2020eednet}. It consists of several processing steps, including noise reduction, white balance, demosaicing, and more. Each step with loss functions in the ISP is usually performed sequentially, the residual error accumulates over the runtime ~\cite{ratnasingam2019deep}. Parameter adjustments in later stages correct the accumulated errors.

Most of the traditional methods use heuristic approaches to derive the solution at each step of the ISP pipeline ~\cite{zhuyu2020eednet}, so numerous parameters need to be adjusted. Moreover, multiple ISP processes executed sequentially with module-based algorithms lead to cumulative errors at each execution step. To minimize those errors, new techniques were researched and, among them, algorithms related to Deep Learning started to get more focus.

\subsection{Deep Learning}
\label{ss.dl}

\wladimir{Even though the research of Machine Learning dates back to the decade of 1950\cite{Rosenblatt58theperceptron}, it was only in the last decade that the advancements in technology have allowed its more complex fields to be extensively explored. The rapid evolution of computational power, coupled with the growing amount of data being produced daily, caused a subtle renewal of interest in the usage of Machine Learning techniques. For that reason, several areas such as Chemistry \cite{Senior2020-gx}, Medicine \cite{Danaee2017-pz}, Economics \cite{Storm2020-jx} and Physics \cite{Askar2019-wh}, have been able to harness its capacities to accelerate or improve their work, directly impacted by this evolution in the field known as Deep Learning. }

\wladimir{Deep Learning, as a subset of Machine Learning, is comprised of algorithms based on Artificial Neural Networks that use several layers of neurons to extract higher level features from the raw data being provided to it \cite{Gomez_co-evolvingrecurrent, Schmidhuber2015-xh, Chen2001-gh}.  This class of algorithms requires a huge amount of computational power that become available only in recent years. In parallel to the high demand of computational power, the capacity of learning of Deep Learning algorithms also rises with the amount of data being provided to the system. For this reason, areas that have a great influx of data in their operation saw in Deep Learning an interesting way to find and understand hidden information.  }

\wladimir{One of the fields that found great results with the use of Deep Learning is the Image Processing (a sub set of Computer Vision), more specifically with the use of Convolutional Neural Networks (CNN). The CNNs are a class of Neural Networks more prone to work with visual imagery for being inspired by biological processes, created in a manner that the connective pattern of the neurons imitates the pattern of the visual cortex of animals \cite{article, Fukushima1980-lk}. Another important aspect of the CNNs is the fact that, following a different path from other networks, they use very minimal pre-processing, being able to learn by themselves to optimize kernels. These features make the use of CNNs more common for Image Processing tasks. }

\subsection{The ISP and Deep Learning Relation}
\label{ss.relation}

The intention of using a CNN to replace the hardware-based ISP is justified by the fact that a CNN can compensate the loss of information in the input images making it more reliable than the traditionally implemented ISP, as traditional ISP is known to accumulate errors at each step \cite{ramanath2005Processing}. ~\cite{ignatov2020replacing} was one of the first to propose a CNN in place of a smartphone ISP camera and provided a RAW-to-RGB dataset using the PyNet network. These demonstrated the potential of CNN for image processing as a replacement for even the most sophisticated ISPs.

Network CNNs not only show significant advantages in low-level vision tasks ~\cite{zhuyu2020eednet}, they also show good results in high-level tasks such as object detection and segmentation ~\cite{chen2017deeplab}. With these advantages, the use of CNNs for transforming RAW images into RGB images became possible.

Despite the good results, there are few works using CNN as a replacement for ISP. ~\cite{ratnasingam2019deep} showed the difficulties in performing the necessary adjustments in traditional ISP pipelines and developed a CNN that performs the ISP pipeline.

\subsection{Comparison with Other Works}
\label{ss.comparison}

One of the great ways to consolidate a growing field in science is to perform surveys about the most current techniques in that field. This way, the access for this kind of information is facilitated, making it easier to understand and choose which technique to use for one's individual case. In the field of Deep Learning, for example, a great number of surveys is already available in many different areas, such as agriculture~\cite{Kamilaris2018}, cyber security~\cite{Berman2019}, autonomous driving~\cite{Grigorescu2020}, medical imaging~\cite{Litjens2017}, and also on more technical areas, such as on CNN ~\cite{Li2020}. But for more recent fields, such as using Deep Learning to replace ~\textbf{ISPs}, it might still be hard to find this kind of gathered information.

While there are already some surveys about individual steps of an ISP, such as demosaicing~\cite{Li2008} and denoising~\cite{Fan2019}, there is still no easy way to find and compare end-to-end ISP deep learning approaches. In this paper we summarize many of these approaches, bringing some of the state-of-the-art Artificial Neural Networks (ANNs) in this field.

\subsection{Scope of this work}
\label{ss.scope}
\jhessica{For this study, the articles were studied according to three main points: }
\begin{itemize}
    \item \textbf{Novelty}: \jhessica{to introduce the most recent and significant works comprising strategies for replacing parts or the entire ISP pipeline through deep learning approaches; and }

    \item \textbf{Recently developed}: \jhessica{all studies considered were published between the years 2019 and 2021, making this study very up-to-date. }
\end{itemize}

\small
\begin{longtable}{c|l}
\caption{Summarization of the approaches considered in the survey.}
\label{tbl:methods}\\
\textbf{Short Name  }                                     & \multicolumn{1}{c}{\textbf{Description}}  \\
\hline
HerNet \cite{mei2019higherresolution}                                            & \multicolumn{1}{p{9.5cm}}{\raggedright Combined CNN's with traditional algorithms to reverse the order of the CFA processing pipeline.} \\
\hline
CameraNet \cite{liang2019cameranet}                                        & \multicolumn{1}{p{9.5cm}}{\raggedright Divides the subtasks with poor correlation, creating a network with two stages.}    \\
\hline
Deep Camera \cite{ratnasingam2019deep}                                      & \multicolumn{1}{p{9.5cm}}{\raggedright Create a network with four parallel paths with convolutional layers and an inverse ISP to synthesize RAW images. }\\
\hline
DRDN \cite{DRDN}                                                &\multicolumn{1}{p{9.5cm}}{\raggedright Color filter array demosaicking based on residual learning and densely connected convolutional neural network.}                \\
\hline
\multicolumn{1}{p{4cm}|}{\centering Deep Demosaicing for Edge Implementation \cite{deepedge} }             & \multicolumn{1}{p{9.5cm}}{\raggedright Discussed the edge implementation of deep learning-based demosaicing algorithms on low-end edge devices}    \\
\hline
\multicolumn{1}{p{4cm}|}{\centering BayerUnify and BayerAug \cite{bayermethods}}                                    & \multicolumn{1}{p{9.5cm}}{\raggedright
Create a method to unify different Bayer patterns and an effective approach for raw image augmentation.}                                      \\
\hline
 VisionISP\cite{visionisp}                                           & \multicolumn{1}{p{9.5cm}}{\raggedright ISP method to increase computer vision applications performance.}                                      \\
\hline
RLDD \cite{guo2020residual}                                     &  \multicolumn{1}{p{9.5cm}}{\raggedright Combined CNN’s with traditional algorithms to reverse the order of demosaicking and denoising. }                                    \\
\hline
DPN \cite{kim2020deep}                                         & \multicolumn{1}{p{9.5cm}}{\raggedright An efficient deep neural network architecture for Quad Bayer CFA demosaicing adopted in submicron sensors.}                                       \\
\hline
CycleISP \cite{Zamir2020CycleISP}                                   &\multicolumn{1}{p{9.5cm}}{\raggedright  Models camera imaging pipeline in forward and reverse directions, producing realistic image pairs for denoising.}                                       \\
\hline
PyNET \cite{ignatov2020replacing}                                & \multicolumn{1}{p{9.5cm}}{\raggedright  Uses a novel pyramidal CNN architecture to replace the mobile camera ISP.   }             \\
\hline
PyNET-CA \cite{kim2020pynetca}                                  &  \multicolumn{1}{p{9.5cm}}{\raggedright Improves PyNET performance by adding channel attention
and subpixel reconstruction modules.  }            \\
\hline
SGNet \cite{9156993}                                             & \multicolumn{1}{p{9.5cm}}{\raggedright Created an adaptative method to treat regions with high and low frequencies.}                                       \\
\hline
\multicolumn{1}{p{4cm}|}{\centering PatchNet and RestoreNet \cite{sun2020learning} }                                    & \multicolumn{1}{p{8cm}}{\raggedright Select the most useful patches from an image for the training step using active learning.}            \\
\hline
AWNet \cite{dai2020awnet}                                        & \multicolumn{1}{p{9.5cm}}{\raggedright Use of wavelet transform and non-local attention mechanism in ISP Pipeline.}                                       \\
\hline
Del-Net \cite{gupta2021delnet}                                     & \multicolumn{1}{p{9.5cm}}{\raggedright  A multi-scale architecture that learns the entire ISP pipeline. Ideal for smartphone deployment.   }              \\
\hline
InvISP \cite{xing2021invertible}                                  &\multicolumn{1}{p{9.5cm}}{\raggedright Reconstruct the RAW data, in addition, to rendering sRGB images using an invertible neural network.    }            \\
\hline
ICDC-Net \cite{uhm2021image}                                        & \multicolumn{1}{p{9.5cm}}{\raggedright An approach with ISP-Net that addresses JPEG image compression in network training.  }                                    \\
\hline
CSANet \cite{ignatov2021learned}                                  &  \multicolumn{1}{p{9.5cm}}{\raggedright Uses cascaded blocks composed of channel attention modules.}                                      \\
\hline
LiteISPNet \cite{zhang2021learning}                                   &  \multicolumn{1}{p{9.5cm}}{\raggedright Method to align pairs of images captured by different cameras during the train.}                                       \\
\hline
TENet \cite{qian2021rethinking}                                  & \multicolumn{1}{p{9.5cm}}{\raggedright Reordered the traditional operation sequence to denoising + super-resolution -> demosaicing.}                                       \\
\hline
ReconfigISP \cite{yu2021reconfigisp}                                   & \multicolumn{1}{p{9.5cm}}{\raggedright Adapted the architecture and parameters according to a specific task.  }            \\
\hline
ISP Distillation \cite{schwartz2021isp}                                     &  \multicolumn{1}{p{9.5cm}}{\raggedright Uses an sRGB image classification model and distills the knowledge of an ISP pipeline.
}            \\
\hline
Merging-ISP  \cite{chaudhari2021mergingisp}                             & \multicolumn{1}{p{9.5cm}}{\raggedright Approach to reconstructing multiple image layers LDR in just one image HDR.} \\
\hline
GCP-Net \cite{Guo2021jdd}                                          & \multicolumn{1}{p{9.5cm}}{\raggedright A joint denoising and demosaicking method for real-world burst images, based on a green channel prior neural network.}                                       \\
\hline
 PIPNet \cite{a2021beyond}                                         & \multicolumn{1}{p{9.5cm}}{\raggedright Uses a deep network to work with joint demosaicing and denoising in CFA patterns.  }              \\
\hline
CURL \cite{moran2021curl}                                       & \multicolumn{1}{p{9.5cm}}{\raggedright Image enhancement method that can be used in the RAW-to-RGB and RGB-to-RAW mapping tasks. }            \\
\hline

\end{longtable}

\begin{figure}[!ht]
    \centering
    \begin{minipage}{0.5\textwidth}
        \centering
        \includegraphics[scale = 0.21]{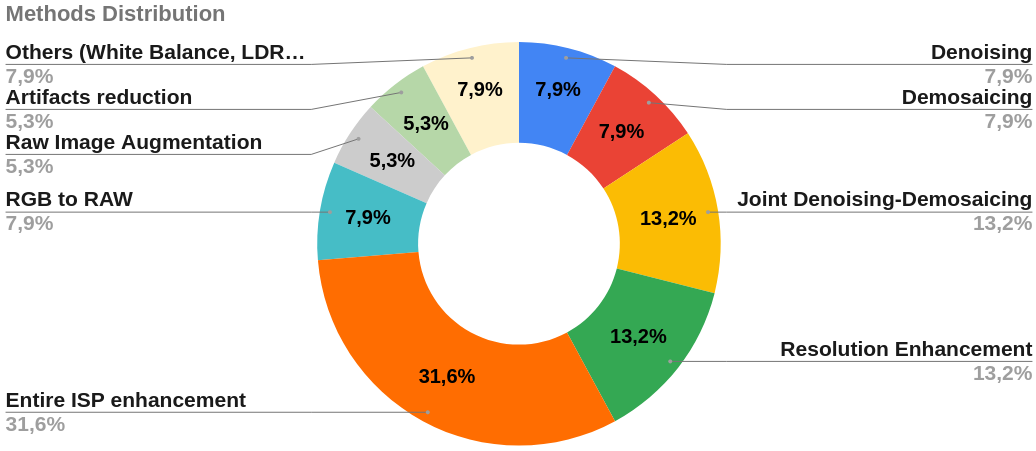}
        \caption{Reviewed ISP tasks distribution}
        \label{f.chart1}
    \end{minipage}\hfill
    \begin{minipage}{0.5\textwidth}
        \centering
        \hspace{1cm}
        \includegraphics[scale = 0.21]{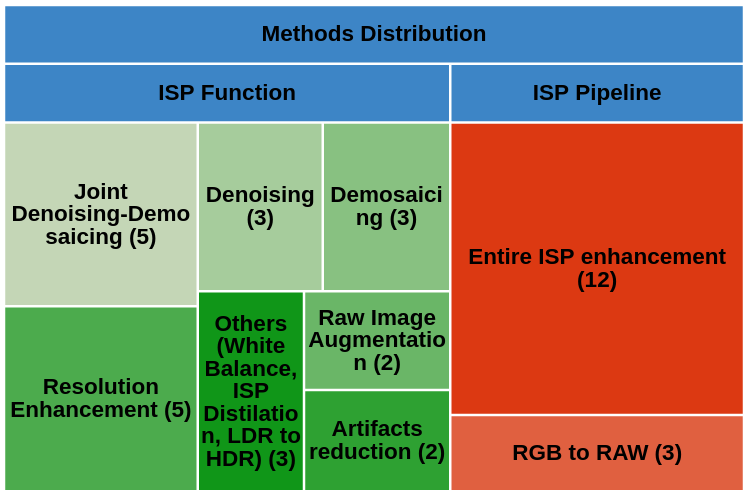}
        \caption{Mapped ISP tasks}
        \label{f.chart2}
    \end{minipage}
\end{figure}

\rodrigo{Those considered papers did not seek the same ISP tasks and improvements. Figure~\ref{f.chart1} shows the ISP studied functions distribution among the reviewed articles and Table~\ref{tbl:methods} gives a list of all methods discussed in this work.}

\rodrigo{Around $30\%$ of the papers proposed an entire ISP pipeline framework using an end-to-end Deep Learning approach. Others developed deep learning solutions for specific stages of an ISP pipeline, such as denoising tasks, joint denoising-demosaicing tasks, resolution enhancement tasks, among others. Some of them also proposed extra and distinct ISP deep learning techniques, like RAW data augmentation and RAW data generation from RGB images by using inversed ISP procedure.}

\rodrigo{Figure~\ref{f.chart2} shows how we mapped those studied ISP tasks. We labeled them into two groups: ISP function, when the proposed solution strikes specifics ISP operations, and ISP pipeline, when the proposed solution settles a RAW to RGB or RGB to RAW operation and an entire ISP procedure.}

\subsection{Work structure}
\label{ss.structure}
\wladimir{The rest of this work is structure as follows: Section 2 describes the overall concept, traditional pipelines and algorithms for ISP. Section 3 contains the resumes for the works collected through this research. Section 4 covers the results of the collected works. Section 5 describes in detail the methodology collected during this research. Section 6 concludes this work.}

\section{Software ISP}
\label{s.introduction}

\brunoiago{Over the last two decades, since the rise in popularity of embedded devices that use digital cameras as a secondary or main feature, the demand for reliable digital image capture and processing systems have grown significantly. Nowadays, processing speed and image quality are great selling points for most of those devices. That being said, studies of image reconstruction systems have never been so important.}

\brunoiago{Traditionally, digital cameras are composed by the bond of two subsystems, the first one being dedicated to the acquisition of signals measured by a grid of photosensitive analog sensors, usually referred to as sensor element~\cite{ramanath2005Processing}. Modern sensor elements have high sensitivity to light variation but are unable to identify color variation on their own. A possible solution to this problem would be to use 3 distinct sensor elements, each with a specific filter to capture a certain frequency range of visible light.}

\begin{figure}[htb!]
  \centering
  \includegraphics[width=0.3\textwidth]{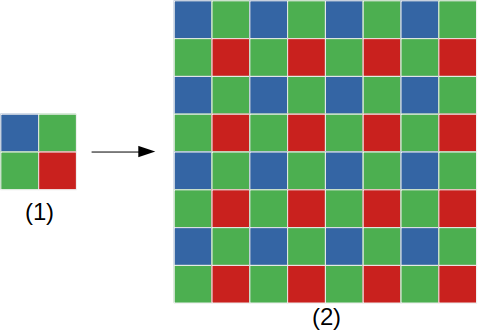}
  \caption{(1) Bayer Pattern and (2) extended pattern. Based on~\cite{WILSON201667}}
  \label{f.bayer}
\end{figure}

\brunoiago{This strategy would bring many technical issues, related primarily to sensor alignment, difference on light incidence, among others, in addition to an increase in hardware cost~\cite{ramanath2005Processing}. Modern sensor elements have a known pattern light frequency filter embedded into them, which makes it possible to reconstruct a color image with a single sensor element. These filters are known as Color Filter Arrays (CFA). The Bayer Color Filter (BCF) is a special type of Red-Green-Blue (RGB) CFA pattern that is widely used in modern image sensors. The BCF is build on the assumption that the human visual system (HVS) has more sensitivity to colors on the green spectrum. Based on that BCF consists of a  2 by 2 grid pattern containing two green, one red and one blue sensor~\cite{ramanath2005Processing} as shown in Figure~\ref{f.bayer}.}

\brunoiago{The BCF filter is placed right in front of the sensor element. The signal resulting from the capture process, filtered by the BCF is called Bayer Array (BA) and is composed of the monochromatic intensity of each pixel, following BCF pattern. RAW image files are composed by BA data in addition to metadata acquired at the time of capture, in this section, information such as capture time, pre-processing strategy, black level, aperture, exposure, ISO, among others, are usually encapsulated in standard Exchangeable Image File (EXIF) data.}

\brunoiago{Modern image sensors, such as OmniVision's OV5647, uses a sensor element composed by a grid of 2624×1956 photosensitive sensors, covered by a layer of BCF. In addition to image data acquisition, this device also provides various processing options such as Automatic Exposure Control (AEC), Automatic White balance (AWB), Automatic Band Filter (ABF), and Automatic Black level Calibration, to provide a RAW output with better overall image quality[4].}

\brunoiago{An ISP pipeline usually referred as the second subsystem of a digital camera. It is build on a series of processes that aim to convert a RAW image file into a visible digital object, in order to display and store the image that was captured. Some of the steps of a traditional pipeline will be explored here in general lines.}

\brunoiago{Traditionally, ISP pipelines are constructed as a sequential series of operations. The Input of an ISP is usually a RAW image and the output is a RGB encoded digital image. Commercially used ISPs may vary in order and type of operations, depending on the manufacturer needs. This information is not available, however, there are some basic operations that are necessary for most ISPs and are used as a basis for the study of this type of subsystem. That said, there are known stages common to almost all traditional ISPs, these stages are shown in the Figure~\ref{f.pipeline} below.}

\begin{figure}[htb!]
  \centering
  \includegraphics[width=0.35\textwidth]{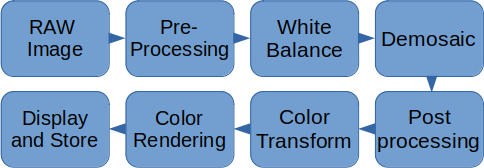}
  \caption{Traditional ISP pipeline. Based on~\cite{karaimer_brown_ECCV_2016}}
  \label{f.pipeline}
\end{figure}

\brunoiago{Although modern image sensors are capable to fix many data acquisition issues[4], a pre-processing  of the data contained in the RAW image is important to verify signal integrity, identify and fix acquisition failures, preventing errors from being propagated through the following operations. Three stages are commonly mentioned in the context of ISP pre-processing: Signal conditioning, defective pixel correction, and black level offset. Signal conditioning refers to normalization, linearization and other operations necessary to adapt the data obtained by the sensor to be processed by the ISP~\cite{ramanath2005Processing}.}

\subsection{Black Level Offset Correction}
\label{ss.black_level}
\brunoiago{Black level offset correction is a necessity created by the imprecision of image sensors, its goal is to correct the ISP input value to reduce black current effects, which tend to increase the light intensity measured by the sensor element and can cause an blur effect in a processed image~\cite{ramanath2005Processing}. The black level offset aims to ensure that the black tones contained in the image are correctly registered. Modern image sensors usually provides an array that contains a mask for black level correction[4].}

\subsection{Defective Pixels Correction}
\label{ss.defective_pixels}
\brunoiago{Defective Pixels are also common and expected acquisition errors up to a certain amount~\cite{9194921}, they occur due to measurement issues caused by production errors, storage methods and temperature problems. The identification of defective pixels is made, in general, from the analysis of the light intensity variation of a central pixel in relation to its neighbors~\cite{9194921}. There are several methods of correction of defective pixels, one of them is to apply the average value of the neighboring points to  the pixel identified as defective~\cite{9194921}.}

\subsection{White Balance}
\label{ss.white_balance}
\brunoiago{After the fixing acquisition issues, a usual first step of a conventional ISP is to perform a white balance on the imputed data. Although the HVS is able to identify the white color of objects illuminated by different types of light sources, digital systems do not have this capability, different frequency range of light results in different measured values~\cite{zapryanov@2012}. White balance is a step that aims to ensure that the measured colors have a natural tone for the human eye after reconstruction~\cite{ramanath2005Processing}. A strategy often used by ISPs is the AWB strategy, while many image sensors already have this feature built in as in OmniVision's OV5647, many ISPs implement it separately. A usual way of doing this is using the gray world assumption, which dictates that the average color of each sampled channel tends to be equal in most cases. Based on this principle, the ratio between the average light intensity measured in the green channel and the others channels used as a basis to make the correction in all pixels~\cite{zapryanov@2012}.}

\subsection{Demosaicing}
\label{ss.demosaicing}
\brunoiago{The most computationally heavy step of a digital image reconstruction refers to the conversion of CFA  data to visible image, this process is called demosaicing. A recurrent strategy to perform this operation is by some variation of weight interpolation of the absolute values of each pixel in the CFA. Although most of the demosaicing techniques used commercially are protected by patent. Some open source applications like RawTherapee are transparent with the demosaicing technique used. In this application, visible image is reconstructed using algorithms such as Adaptive Homogeneity-Directed ~\cite{1395991}.
}

\begin{figure}[htb!]
  \centering
  \includegraphics[width=0.55\textwidth]{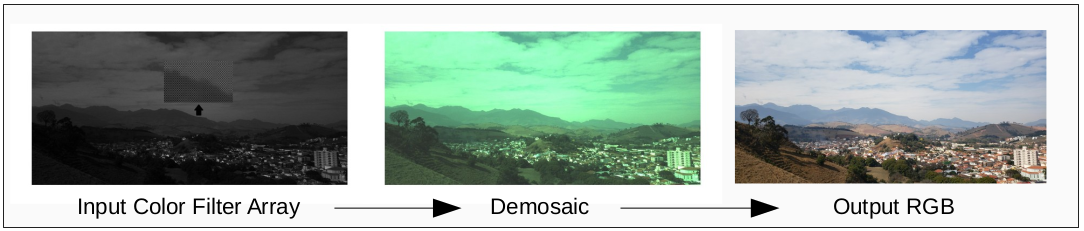}
  \caption{Demosaicing a CFA.}
  \label{f.demosaicing}
\end{figure}

\brunoiago{Figure~\ref{f.demosaicing} shows the comparison between the ISP pipeline output (third image) of the smartphone Samsung G9600 and the RAW image sent by the image sensor (first image). Analyzing the metadata of the sensor output it was possible to identify that the CFA color pattern used in the capture process was the Bayer Green-Red-Blue-Green pattern. This information was used to perform a demosaicing (second image) on the original RAW . The comparison between the second and third images highlights the importance of all steps in an ISP pipeline in order to reconstruct high quality photos.}

\subsection{Denoising}
\label{ss.denoising}
\jhessica{
  Image denoising is a complex step of the digital image reconstruction task, where the goal is to remove the noise from an input image to estimate the original image. This step is usually used in the traditional ISP pipeline because of defects or heterogeneity of the image sensor hardware components, and due to image compression. Image denoising is very important for several applications in the vision computing field and has received a lot of attention over the years~\cite{Motwani2004survey, buades2005review, Sabeenian2012survey, Fan2019, tian2020deep}.  }
  \jhessica{
  Several methods have been proposed for image denoising. They can be classified into classical approaches and deep learning approaches.  The classical approaches encompass the spatial domain method, which applies filters in the image to remove the noise,  and the transform domain method, which changes the domain from the input image and then uses a denoising procedure to improve the image. The deep learning approach, in most cases, is a CNN-based method.  In this survey, some works of deep learning for image denoising will be cited~\cite{anwar2020real, bayermethods, brooks2018unprocessing, 9156993, Hanlin2017Joint, Micha2016Deep}.
}

\subsection{Deblurring}
\label{ss.deblurring}
\matheus{Blur is a general artifact that is hard to avoid in digital image processing and can be caused by various sources like motion blur, out-of-focus, camera shake, extreme light intensity, accumulated error in ISP pipeline, etc. Given this, many handcraft deblurring algorithms exist in the literature to mitigate this problem\cite{Koik2013ALS, Vankawala2015ASO} and some are included in ISPs.  The Razligh and Kehtarnavaz \cite{Razlighi2007Imaeg} proposed a deblurring method for cell phones that takes into consideration the brightness and the contrast to correct the blurred image. On the other hand, Hu et al. \cite{Hu2016ImageDU} consider the use of smartphone's inertial sensors, gyroscope and accelerometer, for kernel estimation and utilize an online calibration to synchronize cameras and sensors. However, many of these classical methods englobe only some cases, realize handcraft feature extraction and are necessary two previous steps\cite{Koik2013ALS}: blur detection and blur classification. Deep learning was an alternative to solving these problems in the last years, possibly jointly all these steps on a unique step\cite{Sahu2019BlindDU}.}

\subsection{Post-processing step}
\label{ss.post_processing}
\brunoiago{Each camera manufacturers can use different, often proprietary, processing methods to improve image quality. The post-processing step aims to make some adjustments to the images that went through the previous processes. Some of the most common post-processing steps used are edge enhancement, removal of colored artifacts and coring~\cite{ramanath2005Processing}. These techniques use heuristics and require considerable fine-tuning.}

\brunoiago{For example, the demosaicing step can introduce artifacts that can be problematic, such as zippered edges and confetti in the rest of the image. In the post-processing stage, it is essential to keep these artifacts to a minimum without losing image sharpness~\cite{ramanath2005Processing}, some camera manufacturers use edge enhancement techniques to make the image more attractive by reducing low-frequency objects contained in the image. The solution to these problems involves many variables, from the size of the capture sensor to the demosaicing technique used.}

\subsection{Rendered Color Spaces}
\label{ss.rendered_color_space}
\brunoiago{Rendered color spaces are generally used as output and have a limited scale, unlike unrendered which are based on scenes. The rendered color space is created based on data extracted from an image in unrendered space and contains a maximum of 8b while the unrendered space has a variable range between 12 and 16b~\cite{ramanath2005Processing}, for this reason, the process of transforming unrendered space to rendered space contains a loss in dynamic range.}

\brunoiago{The most common rendered space is the sRGB~\cite{ramanath2005Processing} color space, which has become common for multimedia. Another common rendering space is the ITU-R BT.709-3, which was created with high-definition televisions in mind. The sRGB standard adopts the primaries defined by ITU-R BT.709-3. It is these patterns that define the methods of transforming unrendered spaces to values of 8b imposed by most output media.}

\brunoiago{Depending on the preview mode, images need to be converted to proper color space. An example is in the case of viewing by a CRT monitor with an additive color system, images need to be converted to an 8b output given the display model used, offset values, color temperature, and values of gamma.}

\brunoiago{When the goal is storage, we have two solutions, professional cameras that have a very large set of sensors and much larger storage space, and generally store the images in a proprietary format or as Tag Image File Format/Electronic Photography (TIFF/ EP). Images stored as TIFF/EP have additional information such as camera settings details and the color transformation matrix~\cite{ramanath2005Processing}. JPEG2000 is an international standard that offers a more efficient compression than the common JPEG standard, in addition to offering several features such as control over data compression size and image resolution. However, despite the benefits that JPEG2000 presents, its computational complexity, and high memory cost are limiting factors.}

\brunoiago{Given the high complexity and need for tuning of modern ISP pipelines, many studies are being made aiming to use machine learning to convert RAW image data into high quality outputs. This work shows state of the art of these studies.}

\section{Know Approaches}

\matheus{In this section, we review the applications of deep learning-based methods to substitute the traditional handcraft ISP pipelines, where include operations as demosaicing, denoising, white balance, tone adjustment, and exposure balance. Also,we briefly summarize some works that use this approach to increase the performance in other computer vision tasks.}

\subsection{HERNet}

\matheus{HighEr-Resolution Network (HERNet) \cite{mei2019higherresolution} is a network that can learn local and global information about high-resolution image patches without excessive consumption of GPU memory. This network has two paths for local and global feature extraction and the introduction of the Pyramid Full-Image Encoder \cite{mei2019higherresolution} that realizes a regularization of the output image and helps to reduce the number of artifacts. Besides, this work proposes training the model with progressively growing the resolution of inputs, which results in performance stability and short training time.}

\matheus{The local information path consists of Multi-Scale Residual Blocks (MSRBs)\cite{Li2018MultiscaleRN} that have two convolutional layers with 3x3 and 5x5 kernels in parallel. Furthermore, in the global information path, was applied an Autoencoder mechanism with modified Residual in Residual (RIR)\cite{zhang2018image} modules for feature extraction. The modified RIRs modules aim to decrease the GPU memory usage, especially to high-resolution images, then the Channel Attention Units were removed, and stacked the remaining. Finally, the authors trained and validated the model with the ZRR dataset\cite{ignatov2020replacing} and used only the L1 loss in training. Unfortunately, the L1 loss favors blurry images in datasets with misalignment.}

\matheus{The local information path consists of Multi-Scale Residual Blocks (MSRBs)\cite{Li2018MultiscaleRN} that have two convolutional layers with 3x3 and 5x5 kernels in parallel. Furthermore, in the global information path, is apply an Autoencoder mechanism with modified Residual in Residual (RIR)\cite{zhang2018image} modules for feature extraction. The modified RIRs modules aim to decrease the GPU memory usage, especially to high-resolution images, then the Channel Attention Units are removed and stacked the remaining. The authors trained and validated the model with the ZRR dataset\cite{ignatov2020replacing} and used only the L1 loss in training. Unfortunately, the L1 loss favors blurry images in datasets with misalignment.}

\matheus{The progressive training was used to train the network, where the input image resolution increased during the training, keeping the same architecture of networks all the time. As a result, this process can make the network converges more quickly. Besides, HERNet won second place in track 1 of fidelity and first place in track 2 of perceptual in AIM 2019 RAW to RGB Mapping Challenge. HERNet won second place in track 1 of fidelity and first place in track 2 of perceptual in AIM 2019 RAW to RGB Mapping Challenge.}

\subsection{CameraNet}

\matheus{The CameraNet\cite {liang2019cameranet} proposes an effective and general framework for a deep learning-based ISP pipeline, with two stages of CNN stacked. The motivation for this is that some subtasks of the ISP pipeline have poor correlation, then the subtasks from the ISP pipeline were divided into two stages: the first stage is the restoration stage with tasks like demosaicing, denoising, and white balance, and the enhancement stage as second stage performing  tasks like exposure adjustment, tone mapping, color enhancement, and contrast adjustment. Besides, two ground truths were created for each image in the datasets HDR+\cite{Hasinoff2016BurstPF} and FiveK\cite{Bychkovsky2011Learning} using Adobe Camera Raw\footnote{https://helpx.adobe.com/br/camera-raw/using/supported-cameras.html} and Adobe Lightroom \footnote{https://www.adobe.com/lightroom}. Each ground truth was used to train a different stage. In the CameraNet pipeline, before these two main stages, the input image is pre-processed with the bad pixels removing, initial demosaicing with interpolation, and converting RGB to CIE XYZ space because it is related to human perception. Moreover, the U-Net is the base model for these two stages, because of the multi-scale extraction features. Some changes were made, like the addition of a fully connected layer in the lowest level of the network and the use of different processing blocks in each stage. While the restoration stage uses plain convolutional blocks, the enhancement stage uses residual connections to details improvement. Furthermore, in the experiments, the CameraNet generated images with less noise, artifacts, better color mapping, and higher qualitative scores than DeepISP\cite{Schwartz2019Deepisp} Network in the HDR+ and, mainly, SID\cite{Chen2018Learning} dataset. The explanation for this difference can be the high level of noise in the SID dataset and the separation of weakly related subtasks in two stages in the CameraNet. In the FiveK dataset, both methods achieve comparables results in SSIM, but the CameraNet obtain superior results in PSNR and Color Error indices because this dataset was captured with high-end cameras, reducing the noise level. DCRAW and CameraRAW, which generate images with default settings, have the lowest results in comparison with Deep Learning methods, explained by the limitation of traditional methods.}

\subsection{Deep Camera}

\matheus{As the first CNN Network proposed to substitute the entire ISP pipeline, the Deep Camera\cite{ratnasingam2019deep} is a small network with four parallel paths: a main path and other three short paths with a convolutional layer in your middle. The explanation for this is the model is very small compared with the ResNet, then the network does not generalize well with the copy of the input to the output of a block. Furthermore, the authors created an inverse ISP to recreate RAW images from a large dataset \cite{Ciurea2003ALI} with 11,000 images and several types of scenes and illuminants, where the training and experiment stages used the resulted dataset.}

\matheus{The CNN model outperformed traditional methods in white balance and image reconstruction tasks, delivering many better images. However, in some images with many different colors, the algorithms are better. Besides, the Deep Camera can do defective pixel correction and be used in other color filter mosaics like the X-Trans color filter by Fujifilm.}

\subsection{DRDN}%

\rodrigo{The DRDN~\cite{DRDN} proposed a convolutional neural network to color filter array demosaicing. Using a mosaiced image as input, the proposed model is trained in an end-to-end manner to generate demosaiced images outputs. Compared to other conventional convolutional neural network-based demosaicing models, the proposed model requires less computational complexity, because it does not require the initial interpolation step for mosaicked input images. It also solved the vanishing-gradient problem experienced by many deep neural networks, due to the residual learning~\cite{resnet} and densely connected convolutional neural network~\cite{Huang2017Densely}. Moreover, the proposed model applied block-wise convolutional neural networks to consider local features and a sub-pixel interpolation layer, generating demosaiced output images more efficiently and accurately.}

\rodrigo{This paper has an exceptional explanation and contextualization about the demosaicing challenging task, citing similar previous methods and highlighting what could be improved in each one. The study detailed the training parameters and aspects related to the considered Datasets. Finally, there is a vast performance comparison, in which the proposed method stood out in the vast majority. On the other hand, the paper did not reveal the inference time obtained during the validation phase, considering that the studied datasets have medium size images. Likewise, the authors could discuss deeper the reason that the DRDN did not reach the best PSNR in some evaluated cases~\cite{DRDN}.}

\subsection{Deep Demosaicing for Edge Implementation}%

\rodrigo{In this paper~\cite{deepedge}, the authors discussed the edge implementation of deep learning-based demosaicing algorithms on low-end edge devices major challenge. They provided an extensive search of deep neural network architectures, obtaining a Pareto front of Color Peak Signal to Noise Ratio (CPSNR) as the loss versus the number of parameters as the model complexity. The article contributed with a valuable reference frame about demosaicing methods, divided into six categories: Edge-sensitive methods, Directional interpolation and decision methods, Frequency domain approaches, Wavelet-based methods, Statistical reconstruction techniques, and Deep learning-based methods~\cite{menon}. Likewise, the authors reviewed other relevant demosaicing aspects, like the unwanted presence of image artifacts, and performance evaluation methods.}

\rodrigo{The study comes up with an exhaustive search of architectures, based on discrete and well-constructed hyper-parameters, like the number of filters and blocks, the skip connections length, and the use of depthwise separable convolutions. It presented a proper methodology and mathematical theory about the neural architecture search and the Pareto building. The highlights were five brand-new theorems in respect to the neural architecture search convergence. Lastly, the designed space with a simple exhaustive search outperformed the state-of-the-art and brought a range of loss versus complexity for edge implementation with varying resource constraints, overcoming drawbacks related to the number of evaluations and the search algorithm complexity. As a drawback, the authors should have discussed more the use of the given architecture search in other image processing tasks, not only in the edge implementation challenge.}

\subsection{BayerUnify and BayerAug}%

\rodrigo{Liu Et al.~\cite{bayermethods} presented two new techniques for DNN-based RAW image denoising. The first is a Bayer pattern unification (BayerUnify) method, that effectively deals with varied Bayer patterns from different data sources. The second method is the Bayer preserving augmentation (BayerAug), allowing proper RAW images augmentation. Associating these two techniques with a modified U-Net, the proposed method achieved a satisfactory SOTA PSNR of 52.11 and an SSIM of 0.9969 in NTIRE 2019 Real Image Denoising Challenge.}

\rodrigo{BayerUnify consists of two stages. In the training phase, the study unified RAW data with different Bayer patterns via cropping. The technique maps the BGGR Bayer format, for example, within the other formats (RGGB, GRBG, and GBRG) and crops the selected area, converting any Bayer pattern to BGGR (or any other chosen pattern). In the testing phase, as the image pixels need to be processed, the technique unified the Bayer patterns via padding. Subsequently, there is the network denoising and the extra pixels removal, disunifying the output images and reversing the pattern conversion.}

\rodrigo{Traditional data augmentation methods are frequently based on image flipping or cropping. However, for RAW images, the flipping procedure may affect the Bayer pattern. BayerAug sorts out this question, combining both flipping and cropping. The paper presented three different flipping methods, which enabled data augmentation on Bayer RAW images without any issues.}

\rodrigo{The study evaluated the proposed method on the Smartphone Image Denoising Dataset (SIDD)~\cite{sidd} - 320 pairs of noisy and noise-free images that covered three different Bayer patterns.  The networks were trained with L1 loss, AdamW~\cite{adamw} optimizer, and 200,000 iterations.}

\rodrigo{The paper presented a satisfactory introduction about the Bayer pattern pre-processing and augmentation, besides discussing related works. The proposed method is properly explained, as well as the network architecture and an exceptional training detailing. At that frame reference, the study shown a promising direction on RAW image processing with deep learning techniques.  As a limitation, the authors could give a comparison to other works in the same NTIRE 2019 challenge, and a deeper discussion about the application of the proposed method in real devices.}

\subsection{VisionISP}%

\rodrigo{Wu Et al.~\cite{visionisp} proposed a particular ISP method for computer vision applications. VisionISP reduced the data transmission needs without relevant information loss, optimizing the subsequent computer vision system performance. The framework consists of three processing blocks. The first block, the Vision Denoiser, reduces the input signal noise and modifies the tuning targets on an existing ISP. The study adopted the technique presented by Nishimura~\cite{nishimura} to optimize the denoising parameters, constructing the denoiser for the computer vision task, not for image quality. The paper also highlighted that the demosaicing step can be skipped and the color filter array image usage, instead of a demosaiced image, did not improve computer vision task performance. The second block, the Vision Local Tone Mapping (VLTM), reduced the bit-depth, achieving similar accuracy with fewer bits per pixel. VLTM used a global non-linear transformation followed by a local detail boosting operator. Lastly, the Trainable Vision Scaler block (TVS) is a generic neural network that processes and downscales the input for a following computer vision engine.}

\rodrigo{VisionISP was trained and evaluated with the KITTI 2D object detection dataset~\cite{kitti}, an autonomous driving benchmark dataset. The study measured the influence of each VisionISP block in the mean average precision(mAP). As a computer vision task sample in the experiments, the authors used the SqueezeDet~\cite{squeezedet} framework and its original code.}

\rodrigo{The paper provided a proper explanation and evaluation of each VisionISP component. As a positive aspect, the experiments show that once TVS is trained, it can be used with other computer vision systems. VisionISP as a whole can be trained jointly or separately with the computer vision backbone. Besides that, each component of the proposed framework enhances a computer vision engine performance and can be deployed independently.}

\rodrigo{Nevertheless, the paper could provide more training details (e.g., epochs number, time inference, hardware, etc.) and a deeper comparison of the VisionISP effectiveness with other computer vision optimization systems at that time.}

\subsection{RLDD}%

\jhessica{
In this paper~\cite{guo2020residual}, the authors combined convolutional neural networks with traditional algorithms to reverse the order of the traditional CFA pipeline (demosaicking and denoising). The method, which we called here RLDD, uses two stages for demosaicking-denoising. The first stage performed the demosaicking by composing the gradient-based threshold-free (GBTF) method~\cite{Pekkucuksen2010Gradient} and a convolutional neural network to overcome the reduction of image resolution in the subsampling operation. The second stage performed the denoising using another convolutional neural network whose goal was to deal with residual noise. The properties of the residual noise were altered due to complex interpolation, and the convolutional neural network aimed to remove it without losing the details of an image.}

\jhessica{
The results were validated on the Kodak~\cite{kodakDataset}, McMaster~\cite{Zhang2011Color}, and Urban 100~\cite{Jia2015Urban} datasets and have shown that this model outperforms state-of-the-art demosaicking and joint demosaicking and denoising algorithms with higher PSNR and SSIM values on all datasets. The results of the visual comparison between the methods confirmed the quantitative values achieved, demonstrating a better visual quality of the images. However, the average running time of demosaicking and denoising of this method did not surpass all the evaluated methods.}

\subsection{DPN}%

\rodrigo{This study proposed a duplex pyramid network (DPN~\cite{kim2020deep}), an efficient deep neural network architecture for Quad Bayer CFA demosaicing adopted in submicron sensors.}

\rodrigo{The article delivers an accurate background in respect to the ISP state-of-the-art challenge and a helpful Quad Bayer CFA Analysis, referencing related Deep Learning based solutions.}

\rodrigo{The proposed architecture consisted of two connected feature map pyramids. One of them is composed of downscaling blocks and the other is composed of upscaling blocks, combined with dense skip connections.}

\rodrigo{As well as the skip connections, inspired by U-Net~\cite{unet}, the given network also applied residual learning, inspired by ResNet~\cite{resnet}. DPN also implemented a Linear Feature Map Growth. Compared to the traditional exponential method, this linear method led to a fewer number of parameters, which is more precise for mobile applications with limited memory constraints.}

\rodrigo{In the results topic, the study brought a comparison against conventional ISP algorithm implemented in Samsung mobile phone, observing improvement in sharpness, color moiré, edges, texture preservation, and visual artifacts reduction. DPN achieved better CPSNR values when compared with other Deep Learning based methods at that frame reference. As a limitation in the proposed network architecture, the input image width and height must be a multiple of $2^{(L+1)}$, where "L" is the resolution level. Otherwise, the input must be cropped before the downscaling blocks. Furthermore, the study could have done tests with larger resolution images which are also captured in mobile phones of that 2019 frame reference, like FullHD and 4K images.}

\subsection{CycleISP}

\rodrigo{Zamir~\cite{Zamir2020CycleISP} Et al. proposed the CycleISP, a framework that models camera imaging pipeline and produces realistic image pairs for denoising both in RAW and sRGB spaces. The authors trained a new image denoising network on synthetic data and achieved state-of-the-art performance on real camera benchmark datasets.}

\rodrigo{CycleISP is a compound of two stages. First, the framework models the camera ISP in forward and reverse directions. Second, it synthesizes realistic noise datasets for the RAW and sRGB images denoising tasks. The CycleISP model introduces the RGB2RAW network branch, the RAW2RGB network branch, an auxiliary color correction network branch, and a noise injection module. The RGB2RAW and RAW2RGB modules are trained independently, followed by a joint fine-tuning.}

\rodrigo{The RGB2RAW branch converts sRGB images to RAW data without requiring any camera parameters. This module network is composed of convolutional layers, proposed recursive residual groups, dual attention blocks, and a final Bayer sampling function, generating a mosaicked RAW output. The RAW2RGB network maps clean RAW images to clean sRGB images. First, the noise injection module is set as 'OFF', followed by a 2x2 block packaging into four channels (RGGB) and an image resolution reduction block. To ensure that the input RAW data may come from any camera and have different Bayer patterns, the RAW2RGB branch applies the Bayer pattern unification technique~\cite{bayermethods}. Next, a convolutional layer and a proposed recursive residual group encode the packed RAW image into a deep feature tensor. Additionally, the authors proposed a color correction branch to the RAW2RGB network, which receives an sRGB image input and generates a color-encoded deep feature tensor.}

\rodrigo{Furthermore, there is a Joint Fine-Tuning that provides optimal-quality images. For this, the RGB2RAW's output becomes the RAW2RGB's input, and the RGB2RAW branch receives gradients from both sub-losses, reconstructing the final sRGB image.}

\rodrigo{To synthesize realistic noise image pairs for denoising in RAW space, the noise injection module is turned 'ON' and includes shot and read noise of different levels to the RGB2RAW's output. After this, CycleISP can generate a clean and its noisy image pair from any sRGB image. For the sRGB space, the CycleISP model is fine-tuned with the SIDD~\cite{sidd} dataset, which contains clean and noisy image pairs in both RAW and sRGB spaces.}

\rodrigo{For the training stage, the authors used the MIT-Adobe FiveK dataset~\cite{Bychkovsky2011Learning}, followed by the Fine-Tuning. They evaluated CycleISP performance with state-of-the-art RAW and sRGB denoising methods, using the DND~\cite{DND} and SIDD~\cite{sidd} benchmarks. CycleISP achieved better results in both scenarios, with almost a 5 times smaller network parameters number than the previous best RAW denoising method~\cite{brooks2018unprocessing} and performance gain against the previous best sRGB denoising algorithm~\cite{anwar2020real}. Besides that, in contrast to the other evaluated models, the proposed method provided clean and artifact-free results, also preserving image details.
The proposed framework modules were appropriately described and reinforced with an ablation study. The authors provide a decent implementation details section, a suitable comparison with related methods, and a solid generalization capability study. As a negative aspect, the paper could show more details at the ablation study, giving more information about individuals contributions of the other CycleISP modules aside from the RAW2RGB branch.}

\subsection{PyNET}%

\jhessica{
PyNET~\cite{ignatov2020replacing} is a novel pyramidal CNN architecture designed to replace the entire ISP pipeline present in smartphones. The proposed method has an inverted pyramidal shape and was composed of five different levels trained from bottom to top where each level is trained sequentially with the trained output being used in the above level training stage. The convolutional filters size in this method varies from 3x3 at level five up to 9x9 at level one.  Therefore, lower levels learn global image manipulation while higher levels learn to reconstruct the final image recovering the missing details at lower levels.  The network is trained using three different loss functions combinations. The lowest levels, four and five, are trained with mean squared error (MSE) to learn global color and brightness correction. Levels two and three are trained by a combination of MSE and VGG-based~\cite{johnson2016perceptual} to refine the color and shape of objects. Finally, level one is trained with MSE, VGG, and SSIM loss~\cite{wang2003multiscale} and performs corrections in the local color processing, noise removal, texture enhancement, etc.   }

\jhessica{
Additionally, the authors present the Zurich RAW to RGB dataset, composed of 20 thousand RAW-RGB image pairs where the RAW images are captured using Huawei P20 smartphone and the RGB images are captured using a  professional high-end Canon 5D Mark IV camera.
}

\jhessica{
To evaluate the method, three experiments are conducted. The first, compared PyNET with the methods SPADE~\cite{park2019semantic}, DPED~\cite{ignatov2017dslr}, U-Net~\cite{unet}, Pix2Pix~\cite{isola2017image}, SRGAN~\cite{ledig2017photorealistic}, VDSR~\cite{kim2016accurate}, and SRCNN~\cite{dong2015image}.  In this comparison, PyNET outperformed all other methods in the PSNR and MS-SSIM metric values.  The second experiment measures the quality of the generated images using the Amazon Mechanical Turk\footnote{https://www.mturk.com} platform, compared the images of PyNET,  Visualized RAW, and Huawei P20 ISP with the images produced by the Canon 5D Mark IV DSLR camera. In this comparison, the image produced by PyNET reached the better MOS result in comparison to the target DSLR camera. Finally, the last experiment tested the PyNET using RAW images captured by the BlackBerry KeyOne smartphone without re-training the network. The image produced by PyNET was compared with the image produced by BlackBerry’s ISP, and PyNET generated good results, however, the results were not ideal in terms of exposure and sharpness. }

\subsection{SGNet}%

\matheus{Many methods have joined highly correlated tasks have success, decreasing the accumulated error in the several process units in the ISP pipeline. Thus, the SGNet\cite{9156993} joins demosaicing and denoising in a unique network.}

\matheus{As the correction of high-frequency regions in images is more complicated, the authors propose extracting a density map representing the frequency of areas of the picture. This density map can help the network know each region's difficulty level and adapt better than other models in areas with high frequency. Furthermore, half of the Bayer pattern comprises green pixels; subsequently, it is easier to recover the missing pixels from this channel. For this reason, the network has a branch to reconstruct the green channel, where consequently helps reconstruct other channels. Besides, SGNet uses the Residual-in-Residual Dense Block (RRDB) to feature extraction in both branches. Additionally, this network was trained in a set of loss functions, which consider the reconstruction fidelity of the green channel, full image, the objects, textures edges, and noise removal.}

\matheus{SGNet outperforms state-of-the-art methods in terms of PSNR and SSIM in datasets aimed at the super-resolution, denoising, and demosaicing tasks. Furthermore, compared with the ADMN\cite{Hanlin2017Joint}, CDM\cite{Tan2017COLORID}, Kokkinos\cite{kokkinos2018deep}, Deepjoint\footnotemark~\cite{Micha2016Deep}, and FlexISP\cite{Felix2014Flexisp}, the SGNet can remove moiré artifacts more effectively and give images with more definition in high-frequency areas than ADMN and Deepjoint. However, as a negative point, this work did not show a preoccupation with computational efficiency, which is essential to applications that use demosaicing and denoising.}

\subsection{PatchNet and RestoreNet}
\jhessica{In this paper~\cite{sun2020learning}, the authors proposed a method based on active learning (data-driven)  that learns to select the most suitable patches from an image for the training step without adding additional cost to the inference step.  To do this, the method, called PatchNet, assigns a weight to each patch that defines whether it will be used or ignored during training.  This method is a feed-forward network with multiple stages where each stage is composed of several convolutions blocks and a down-sampling operator. Then gradually the stages transform the image into a set of trainability scalars that are finally binarized to obtain the network output.  }

\jhessica{In addition to PatchNet, the authors also proposed RestoreNet, an architecture that applies the structural knowledge extracted from PatchNet and is responsible for restoring the original image.  }

\jhessica{The results were validated on the Vimeo-90k~\cite{Xue2019}, MIT Moire\cite{Micha2016Deep}, and Urban 100~\cite{Jia2015Urban} datasets and compared with the Kokkinos~\cite{kokkinos2018deep}, SGNet~\cite{9156993}, CDM~\cite{Tan2017COLORID}, and DeepJDD\footnotemark[\value{footnote}]~\cite{Micha2016Deep} methods. The methods were compared at three different noise levels (5, 15, 25). In all comparisons, the proposed method achieved better PSRN values in the JDD task.}

\jhessica{The ablation studies analyzed the effects of different patch sizes when PatchNet is evaluated on Demosaicing and shown that performance improves as patch size increases.  It would be interesting to study the computational costs involved in increasing patch sizes and how much this would change the network's complexity.  The study continued with experiments of PatchNet on JDD and compared it with the methods mentioned above. Only the PSNR metric was used for comparison, but it would have been interesting to evaluate other metrics. }

\subsection{AWNet}
\wladimir{In this work~\cite{dai2020awnet}, the researchers propose a method capable of enhancing smartphone generated images by replacing the base ISP by a U-Net~\cite{unet} resembled CNN (Convolutional Neural Network), called AWNet.}

\wladimir{The network is divided into two branches, each using different inputs and, thus, different models. The first branch, using the RAW model, receives 224 x 244 x 4 RAW images and the second branch, using the demosaiced model, receives 448 x 448 x 3 demosaiced images. Both branches are trained separetely and the results are averaged during test.}

\wladimir{The structure of this network, following the U-Net, applies three main modules to the inputs, for each branch: global context res-dense, residual wavelet up-sampling and residual wavelet down-sampling. The res-dense module is applied to extract the low frequency components after the discrete wavelet transform (DWT), which are sent to the layer below, while the down-sampling extract all the components. After the extraction, both sets of components for each layer are up-sampled and concatenated with the layer above. Finally, a pyramid pooling module is applied, generating the output for that branch.}

\wladimir{For the test phase, a self-ensemble mechanism was applied, made up of 8 ensemble variants. Those variants where, then, evaluated using the PSNR (dB) values, which would be used as weights to generate the final predictions of the model. The chosen PSNR were 21.36 dB for the RAW model and 21.52 dB for the demosaiced model. By applying this tunned model to the two tracks of the Zurich dataset from AIM 2020 Learned Smartphone ISP Challenge~\cite{ignatov2020aim}, the study reached the 5th and 2nd positions, respectivelly.}

\wladimir{Using the results as a justification for the use of the wavelet transform and the global context blocks, the researchers compared the results of AWNet against other popular network architectures, like U-Net, RCAN~\cite{zhang2018image} and PyNet~\cite{ignatov2020replacing}, with the use of the ZRR dataset. By comparing their performance, the researches found that AWNet is able to outperform U-Net, RCAN and the current state of the art, PyNet.}

\subsection{PyNET-CA}%

\jhessica{PyNET-CA is an end-to-end mobile ISP deep learning algorithm for RAW to RGB reconstruction~\cite{kim2020pynetca}. This network improves PyNET~\cite{ignatov2020replacing} performance by adding channel attention and subpixel reconstruction modules and decreasing training time. PyNET-CA has an invertible pyramidal structure for considering the local and global features of the image. The Basic modules of PyNET-CA are the channel attention module based on~\cite{zhang2018image}, the DoubleConv module, which has two operations of 2D convolution with a LeakyReLU activation, and the MultiConv channel attention module, which concatenates the features from the DoubleConv modules and a channel attention module.}

\jhessica{The superpixel reconstruction module helps the network reconstruct the final image with quality and better computational efficiency. For this, PyNET-CA upsamples the image with the MultiConv channel attention module, followed by a 1×1 convolution layer and upsamples the features by subpixel shuffling at the final level of the model. The results were presented on Zurich Dataset~\cite{ignatov2020replacing} where it has shown better PSNR and SSIM values when compared to PyNET~\cite{ignatov2020replacing}. This paper did not present the number of network parameters and although the authors cite the decrease in training time, a table comparing these results to PyNET~\cite{ignatov2020replacing} was not presented. }

\subsection{Del-Net}

\jhessica{Del-Net~\cite{gupta2021delnet} is a single-stage end-to-end deep learning model that learns the entire ISP pipeline to convert RAW Bayer data to high-quality sRGB-image. This network uses a combination of Spatial and Channel Attention blocks (modified UNet)~\cite{zamir2020learning} and Enhancement Attention Modules blocks~\cite{anwar2020real}. The Spatial and Channel Attention blocks allow the network to capture global details both spatial-wise and channel-wise, therefore helping with color enhancement. The Enhancement Attention Modules blocks help in denoising, which improves the PSNR value. The images generated by Del-Net are visually comparable to the state-of-the-art networks (PyNET~\cite{ignatov2020replacing}, AWNet~\cite{dai2020awnet}, and MW-ISPNet~\cite{ignatov2020aim}) when considering the color enhancement, denoising, and detail retention capabilities while presenting a reduction in Mult-Adds (Number of composite multiply-accumulate operations for an image). Altogether, this makes the network ideal for smartphone deployment. It also has a competitive trade-off between accuracy metrics and complexity. The results were presented on Zurich Dataset~\cite{ignatov2020replacing} where it has shown better detail retention compared to PyNET~\cite{ignatov2020replacing}, better denoising compared to MW-ISPNet~{ignatov2020aim}, and better colour enhancement compared to AWNet~\cite{dai2020awnet}. Although, the detail recovery capability of Del-Net is inferior to that of MW-ISPNet~\cite{ignatov2020aim}.}

\subsection{InvISP}

\jhessica{InvISP~\cite{xing2021invertible} redesigns the ISP pipeline allowing the reconstruction of RAW images almost identical to camera RAW images without any memory overhead and also generates human-pleasing sRGB images like traditional ISPs. This is interesting since end users can only access processed sRGB images because RAW images are too large to store on devices. The reconstruction of RAW images in this method is done by the compression of RGB images with the inverse process. To achieve this goal, the authors designed a RAW-to-RGB and RGB-to-RAW mapping from an invertible neural network consisting of a stack of affine coupling layers and an invertible 1x1 convolution. In addition, a differentiable JPEG compression simulator was integrated into the model, allowing the reconstruction of near-perfect RAW images from JPEG images by Fourier series expansion. The network was trained bidirectionally to jointly optimize the RGB and RAW reconstruction process. Model evaluation was performed on the Canon EOS 5D subset and Nikon D700 subset from the MIT-Adobe FiveK dataset~\cite{Bychkovsky2011Learning}. To render the ground truth of sRGB images from RAW images, the LibRAW library was used, which allows simulation of the steps of an ISP pipeline. The experiments demonstrated an improvement of PSNR over the RAW synthesizing methods UPI ~\cite{brooks2018unprocessing} and CycleISP~\cite{Zamir2020CycleISP}, which implies a more accurate retrieval of RAW images. The method was compared to Invertible Grayscale~\cite{acm2019tog} and U-net~\cite{Chen2018Learning} baselines and the results showed better PSNR and SSIM values, indicating a more robust model for RAW image retrieval and RGB image generation. }

\subsection{ICDC-Net}

\jhessica{In this paper, the authors have proposed an ISP-Net that addresses JPEG image compression in network training ~\cite{uhm2021image}, which we called here ICDC-Net. The fact that images can lose information in the compression process has not been addressed on previously ISP pipelines with convolutional neural networks.}

\jhessica{To this end, the authors applied a fully convolutional compression artifacts simulation network (CAS-Net)}. \jhessica{This network can add JPEG compression artifacts to an image, and it is trained by inverting the inputs and outputs needed for training compression artifact reduction networks. In this work, the authors connected the CAS-Net to an ISP network, so the ISP network can be trained with consideration to image compression, taking compression artifacts into account. The ISP-Net used in this work was U-Net with channel attention module~\cite{uhm2019wnet} and the architecture of CAS-Net was U-Net~\cite{unet} without channel attention module.}

\jhessica{Results are present on the Nikon D700 subset from the MIT-Adobe FiveK dataset~\cite{Bychkovsky2011Learning}. To render the ground truth of sRGB images from RAW images, the LibRaw library was used. The sRGB images were compressed with two different QFs, 80 and 90, and each model was trained separately. The experimental results have shown that this proposed network can produce better quality images when compared to its compression agnostic version.}

\subsection{CSANet}

\matheus{As the second place in the Mobile AI 2021 Learned Smartphone ISP Challenge\cite{ignatov2021learned} and first place in PSNR score, the Channel Spatial Attention Network (CSANet)\cite{csanet} is a network that aims at computational performance and the quality of final image results, inferred at most 90.8 ms per image. The network has three main parts: downscale, processing blocks cascaded, and upscale. In the first part, to reduce the computational time and number of parameters, the authors did a downscale with a strided convolution block following a conventional convolution to feature extraction. Sequentially, the network has a Double Attention Module (DAM) inspired by the Convolutional Block Attention Module (CBAM)\cite{woo2018cbam}. The DAM comprises a spatial attention module to learn spatial dependencies in the feature maps and a channel attention module to learn the inter-channel relationship of features maps. And in the last part has the convolution transpose and depth to space to upscale to a final RGB image. Another essential part of this work is the loss function composed by the Charbonnier loss\cite{zhang2018image}, the SSIM loss, and the Perceptual loss to decrease the perceptual difference between the generated image and the ground truth.}

\matheus{In quantitative metrics on the validation dataset of Mobile AI 2021 Learned Smartphone ISP Challenge, the CSANet outperforms the PUNet, the baseline model for this Challenge, and has a shorter runtime. Also, it was comparable results and better inference time than the AWNet \cite{dai2020awnet}, a network with a high score in the AIM 2020. Compared with other candidates in the Challenge, this work was ranked second with satisfactory runtime e highest quality score. The CSANet can be used in embedded systems as the tests in smartphones showed. Besides, as expected, the use of the modules of attention helped in better color mapping.}

\subsection{LiteISPNet}

\matheus{In some datasets, the RAW and RGB images are captured with different cameras. Consequently, the pairs of images have misalignment and color inconsistency, difficulting the training process and producing blurry results. Thinking about this problem, Zhang et al. \cite{zhang2021learning} proposed a method to train the networks with misaligned images and map RAW to RGB. The authors used the pre-trained optical flow estimation network, PWC-Net \cite{sun2018pwcnet}, to align the image pairs and designed a global color mapping (GCM) to match the color between the input and 'target images to facilitate alignment. Besides, the LiteISPNet is responsible for mapping the RAW-to-RGB. It simplifies MW-ISPNet \cite{ignatov2020aim}, which proposed a U-Net based multi-level wavelet ISP network, reducing the number of RCAB in each residual group and changing the position of convolutional layer and residual group before each wavelet decomposition. These changes decreased the model size and running time by approximately 40\% and 20\%, respectively.}

\matheus{The authors tested the network in two datasets, the ZRR dataset\cite{ignatov2020replacing}, and the SR-RAW\cite{zhang2019zoom}, with two variants of ground truth: the original GT and align GT. In the ZRR dataset, the LiteISPNet was compared with three states of the art (PyNet\cite{ignatov2020replacing}, AWNet\cite{dai2020awnet}, and MW-ISPNet) and outperformed all metrics on the aligned GT but was a little worse than MW-ISPNet in the SSIM metric on the original GT. Moreover, the GAN version of this model obtains better perceptual results in the LPIPS metric\cite{zhang2018unreasonable}. Finally, in qualitative comparison, the network could retain more fine details than other models. With the SR-RAW dataset, the authors also compared with SR methods, as SRGAN\cite{ledig2017photorealistic}, ESRGAN \cite{wang2018esrgan}, SPSR \cite{ma2020structurepreserving}, and RealSR\cite{Ji2020RealWorldSV}. It generates images with less noise, less blurry, more details, and it had better scores in almost all metrics, losses only in the PSNR metric on original GT, which favors blurred images.}

\matheus{In conclusion, this work outperformed state-of-the-art models in ISP and SR problems, in addition to provide a new method to train DNN models with misalignment datasets. Besides, this new method enabled a lightweight network, such as used at work, and generated results near or higher than more robust models. However, the authors did not test the model in embedded systems.}

\subsection{TENet}

\matheus{Usually, the ISP pipeline is an operations sequence with three core components in a fixed order: demosaicing, denoising, and super-resolution. However, Qian et al. \cite{qian2021rethinking}, in extensive experiments, shown that a simple reordering of the operation sequence can increase the image quality. Then, the authors created the Trinity Enhancement Network (TENet), a network that reordered the operation sequence to denoising(DN), super-resolution(SR), and demosaicing(DM). The DN block is the first because the noise on a RAW image has a Gaussian-Poisson distribution \cite{Hasinoff2014}, then more straightforward to resolve; the RAW image noise can hinder subsequent tasks; also, this noise become complex over image processes operations. Furthermore, the DM in higher resolution images results in fewer artifacts, and super-resolution algorithms could amplify the artifacts generated by DM. Therefore the SR was the second block in this architecture.}

\matheus{As the DN produces blur in the image, the authors joined the DN and SR in a unique block eliminating the accumulated error over image operations and resulting in this final pipeline: DN + SR -> DM. To effectively leverage in consideration these two stages, the loss function is composed of two losses: the $\Lb_{joint}$, which is the $l_2$-norm loss on the final output image, and LSR, the $l_2$-norm loss between the DN+SR result and the high-resolution noise-free mosaiced image of the input image. Thus, the final loss was the sum of these two losses. Besides, the authors used the Residual in Residual Dense Block(RRDB) \cite{wang2018esrgan} to construct the central part of all blocks.}

\matheus{They also notice that previous datasets that synthesized the DM are sub-optimal for three reasons: 1) The images used to synthesize RAW images are the result of interpolation by the camera ISP; 2) the model was trained to learn an average DM algorithm used in the camera ISP; 3) The synthesized RAW images had less information than real RAW images. For this reason the PixelShift200 Dataset cite{qian2021rethinking} was created with 200 2k-resolution full color sampled images. Each pixel of images was all color information without demosaicing because of the pixel shift technique. Besides, from these high-resolution RAW images was created the low-resolution RAW images through the bicubic downsampling kernel \cite{dong2015image}, mosaic kernel \cite{brooks2018unprocessing}, and addition of the Gaussian-Poisson noise model \cite{Hasinoff2014}.}

\matheus{The model was compared with the ADMM \cite{Hanlin2017Joint}, Condat \cite{Laurent2012Joint}, Flex-ISP \cite{Felix2014Flexisp}, and DemosaicNet\footnotemark[\value{footnote}]~\cite{Micha2016Deep} on the commonly used benchmark datasets to denoise and demosaicing tasks: Urban 100 \cite{Jia2015Urban}, Kodak, McMaster \cite{Zhang2011Color} and BSD100 \cite{David2001Database}. TENet outperforms these models in qualitative metrics, therefore generating much less moiré, color artifacts, and more fine-grained textures. The network generated clean images with accurate details validated in the datasets with the addition of Gaussian white noise, where the ADMM and DemosaicNet generate smooth results, FlexISP does not treat the noise correctly, and the ADMM generates color aliasing artifacts.}

\subsection{ReconfigISP}%

\jhessica{
ReconfigISP~\cite{yu2021reconfigisp} is a reconfigurable ISP where the architecture and parameters are adapted according to a specific task. To accomplish this, the authors have implemented several ISP modules and given a specific task, an optimal pipeline is configured by automatically adjusting hundreds of parameters. This method maintained the modularity of the steps in an image reconstruction process, where each module performs a clear role in the ISP pipeline and allows back-propagation for each module by training a differentiable proxy. The differentiable proxy aimed to imitate a non-differentiable module via a convolutional neural network, thus allowing the optimization of the module's parameters. Therefore, the ISP architecture was explored with neural architecture search, where modules receive an architecture weight and are removed if the weight is below a pre-set threshold. This also reduces computational complexity and speeds up the training process. The loss function in this network was chosen according to the specific task desired.}

\jhessica{
To validate the effectiveness of this proposal, the authors performed experiments with image restoration and object detection with different sensors, light conditions, and efficiency constraints. The results were validated over the SID Dataset~\cite{Chen2018Learning} and S7 ISP Dataset~\cite{Schwartz2019Deepisp} and showed that this network outperforms the traditional ISP pipelines achieving a higher PSNR value than Camera ISP. When compared to U-Net~\cite{Chen2018Learning}, the method obtained a better PSNR value for data with smaller numbers of patches in training but a lower value for larger-scale data.}

\subsection{ISP Distillation}%

\jhessica{
In ISP Distillation~\cite{schwartz2021isp}, the authors proposed a model for image classification with RAW images using an sRGB image classification model and Knowledge Distillation~\cite{hinton2015distilling} of an ISP pipeline to reduce the compute cost of the traditional ISP. Traditional ISP pipelines focus on human vision, while this paper provided a solution for machine vision only. Because of this, the authors applied the vision models directly to the RAW data.}

\jhessica{A dataset of RAW and RGB pairs was used to overcome the performance drop that occurs when data was trained directly on RAW images. This dataset is used to pre-train a model that was subsequently distilled to another model responsible for treating directly the RAW data.}

\jhessica{To validate the proposal, two cases were tested. The first was by discarding denoising and demosaicing pre-processing on a model pre-trained on the ImageNet dataset~\cite{russakovsky2015imagenet}. The second was to discard the entire ISP pipeline in a model pre-trained on the HDR+ dataset ~\cite{Hasinoff2016BurstPF}.}

\jhessica{ResNet18~\cite{resnet} and MobileNetV2~\cite{sandler2019mobilenetv2} were used for validation. Both experiments demonstrated good performance when evaluated on top-1 and top-5 metrics. Therefore, ISP Distillation is a step towards achieving similar classification performance on RAW images when compared to RGB.  Although the paper cited that ISP Distillation saves the computation cost of the ISP, the computation cost was not presented in the article. }

\subsection{Merging-ISP}

\guilhermeHepfener{Merging-ISP~\cite{chaudhari2021mergingisp} consists of a deep neural network responsible for reconstructing multiple image layers LDR (low dynamic range) in just one image HDR (high dynamic range). Thus, the input data contains RAW images in dynamic or static scenes, where the network maps them and convolute all the layers in one exit HDR. Before the convolution, a DnCNN~\cite{zhang2017beyond} conception applies a filter with 5x5 size and 64 layers, then applies two filters, both with 5x5 size and 64 layers, and, finally, applies three filters with 1x1 size and activation sigmoid. The output obtained, the data volume is reduced without applying another trainee and the LDR merge in just one HDR, comprising four convolutional layers Merging-ISP: Multi-Exposure High Dynamic Range Image Signal Processing 7 with decreasing receptive fields of 7 × 7 for 100 filters in the first layer to 1 × 1 for three filters in the last layer. Note that it was not necessary to apply an optical flow on input data.}

\guilhermeHepfener{To train the network, synthetic and real datasets based on Kalantari~\cite{kalantari2017deep} datasets were used. The data contained dynamic and statics scenes, how was stated before. Secondly, rotation techniques were used to increase the dataset, contributing to extracting 210000 non-overlapping patches of size 50 × 50 pixels using a stride of 50. Besides, they perform training over 70 epochs with a constant learning rate of 10e$^{-4}$ and batches of size 32. During each epoch, all batches are randomly shuffled. }

\guilhermeHepfener{In comparison to other merging-ISP methods, this one approach obtained the best result in PSNR, SSIM and HDR-VDP-2 parameters.    }

\subsection{GCP-Net}

\rodrigo{Guo Et al. ~\cite{Guo2021jdd} studied a CNN-based Joint Denoising and Demosaicing method for real-world burst images. For this task, since the green channel has twice the sampling rate and better quality than the red and blue channels in CFA RAW data, the authors proposed a green channel prior neural network - the GCP Net. This model extracted the GCP features from green channels to conduct the deep feature modeling, upsampling the image and evaluate the frames offset, relieving the noise impact. The given work also sought out realistic noise models~\cite{Mildenhall2018}, ~\cite{brooks2018unprocessing}, and a set of burst images instead of a single CFA image.}

\rodrigo{The GCP-Net structure is composed of two branches - a GCP branch and a reconstruction branch. By using several convolutional and LReLu blocks~\cite{xu2015empirical}, the GCP branch extracts the green features from the noisy green channels concatenation and their noise level map.}

\rodrigo{The reconstruction branch estimates the clean full-color image. The branch consists of three blocks - the intra-frame module (IntraF), the inter-frame module (InterF), and the merge module - and utilizes the burst images, the noise maps, and the GCP features as the guided information. The IntraF block models the deep features of each frame and guides the feature extraction using the GCP features. The InterF uses a deformable convolution~\cite{Dai2017} in the feature domain to make up for the shift between frames. A pyramidal processing is applied to handle possible large motions, just like EDVR~\cite{Wang2019} and RViDeNet~\cite{Yue2020}. Furthermore, InterF includes an LSTM regularization in the offset estimation, providing the temporal constraint. The merge module provides adaptive upsampling for the full-resolution image reconstruction.}

\rodrigo{The study synthesized trained data using the Vimeo-90k open high-quality video dataset~\cite{Xue2019}. The training lasted two days, using the PyTorch framework and two Nvidia GeForce RTX 2080 Ti GPU. The authors used the Vid64~\cite{Liu2014} and the REDS4~\cite{Wang2019} datasets for the ablation experiments.}

\rodrigo{For the comparison experiments, the authors tested the proposed model on synthetic data and real-world data. In both scenarios, the GCP-Net achieved superior quantitative and qualitative performance to other state-of-the-art Join Denoising-Demosaicing algorithms, such as FlexISP~\cite{Felix2014Flexisp} and ADMM~\cite{Hanlin2017Joint}. The Paper provided a complete introduction and related work explanation. There was also a proper and detailed experiment section, including fine points about training parameters. The ablation study validated the effectiveness of major GCP-Net components. The chosen comparison datasets enhanced the proposed model value, especially with real-world data verification.}

\subsection{PIPNet}

\jhessica{In this paper~\cite{a2021beyond}, the authors proposed a deep network to work with joint demosaicing and denoising in Quad Bayer CFA and Bayer CFA patterns. The proposed network uses attention mechanisms and is oriented by an objective function, including news perceptual losses to produce pleasure images on a pixel-bin image sensor. This network, defined as a pixel-bin image processing network (PIPNet), uses UNet as a framework and traverses different feature depths through downscaling and upscaling operations to leverage the architecture used. The authors also extended the method to reconstruct and enhance perceptual images captured with a smartphone camera.   }

\jhessica{The results were validated on the MSR demosaicing dataset~\cite{khashabi2014joint}, BSD100~\cite{David2001Database}, McMaster~\cite{Zhang2011Color}, Urban 100~\cite{Jia2015Urban}, Kodak~\cite{kodakDataset}, and WED~\cite{ma2016waterloo} datasets and compared with Deepjoint\footnotemark[\value{footnote}]~\cite{Micha2016Deep}, Kokkinos~\cite{kokkinos2018deep}, Dong~\cite{dong2018joint}, DeepISP~\cite{Schwartz2019Deepisp}, and DPN~\cite{kim2020deep} methods at three different noise levels (5, 15, 25). In all comparisons, PIPNet performed better over the PSNR, SSIM, and DeltaE2000 metrics. Qualitatively, the method also outperformed the other approaches. However, the network was tested only with data collected by traditional Bayer sensors, which may hinder network performance in other scenarios. }

\footnotetext{Method from paper Deep Joint Demosaicking and Denoising~\cite{Micha2016Deep}, called by~\cite{9156993, a2021beyond} as Deepjoint, by~\cite{qian2021rethinking} as DemosaicNet, and by~\cite{sun2020learning} as DeepJDD.}

\subsection{CURL}

\jhessica{In this paper~\cite{moran2021curl}, the authors proposed a method for enhancing an image inspired by photographers who perform image retouching based on global image adjustment curves.  This method, called CURL, can be used in two different scenarios. The first is the RGB-to-RGB mapping where an input RGB image is mapped to another visually pleasing RGB image and the second scenario is the RAW-to-RGB mapping where the entire ISP pipeline is done.}

\jhessica{This method is composed of two architectures called Transformed Encoder-Decoder (TED) backbone and CURL block. The TED is similar to U-Net~\cite{unet} but without the skip connections, except for the level-1 skip connections which were replaced by a multi-scale neural processing block that provides enhanced images via local pixel processing to the CURL block. The CURL block is a Neural Curve Layers block that exploits the representation of the image in three color spaces (CIELab, HSV, RGB) intending to globally refine its properties through color, luminance, and saturation adjustments guided by a new multi-color space loss function. The CURL loss function aims to optimize the final image in its different properties such as chrominance, hue, luminance, and saturation.  }

\jhessica{Two experiments were done for the validation of the method, where the medium-to-medium exposure RAW to RGB mapping and the predicting the retouching of photographers for RGB to RGB mapping was evaluated.  In the first, results were validated on the Samsung S7 dataset~\cite{Schwartz2019Deepisp}, and CURL scored the best PSNR and LPIPS metrics when compared to the U-Net~\cite{unet} and DeepISP~\cite{Schwartz2019Deepisp} methods, but tied with the DeepISP method on the SSIM metric.  In the second, results were validated over the MIT-Adobe FiveK dataset~\cite{Bychkovsky2011Learning} and compared with HDRNet~\cite{gharbi2017deep}, DPE~\cite{chen2018deep}, White-Box~\cite{hu2018exposure}, Distort-and-Recover~\cite{park2018distort}, and DeepUPE~\cite{wang2019underexposed} methods where CURL scored better on PSNR and LPIPS metrics, but DeepUPE scored the best on SSIM.  Qualitatively CURL generates images very pleasing to the human eye. }

\section{Methodology}
\label{s.methodology}

\matheus{This section describes how we made a qualitative comparison among the works covered in this survey and the analysis of these papers to highlight points of improvement, highlights, and ways to evolve this field of study.}

\subsection{Datasets}
\label{ss.datasets}

\matheus{For the quantitative evaluation, we provide a comparison among the works covered in this survey with the more explored datasets. To do this, we use the results provided by these works and the most commonly used metrics in the image restoration task, PSRN and SSIM, as the base for comparison. We provide a brief discussion on each most commonly used dataset on the studies we analyzed. } \jhessica{Table~\ref{tbl:dtinfo} gives a list of all datasets discussed in this section with details about the number of images, size, and link to download. }

\begin{table*}[htb!]
	\small
	\centering
	\caption{Summarization of datasets considered in the survey and their respective amount of images, size, and link to download.}
	
	\begin{tabular}{c|c|c|c}
		\textbf{Dataset name}    & \textbf{Nº images} & \textbf{Size} & \textbf{Link to download}  \\\hline
		Zurich RAW to RGB \cite{Ignatov2019AIM2C}& 20.000          & $\approx$ 22 GB & \multicolumn{1}{p{5cm}}{\raggedright \url{http://people.ee.ethz.ch/~ihnatova/pynet.html}} \\\hline
		Urban 100 \cite{Jia2015Urban}        &   100    &          1.14 GB & \multicolumn{1}{p{5cm}}{\raggedright \url{https://github.com/jbhuang0604/SelfExSR}}         \\\hline
		McMaster dataset \cite{Zhang2011Color} & 18 & 13.6 MB & \multicolumn{1}{p{5cm}}{\raggedright \url{https://web.comp.polyu.edu.hk/cslzhang/CDM_Dataset.htm}} \\\hline
		Kodak & 25  & 119.7 MB   &   \multicolumn{1}{p{5cm}}{\raggedright \url{http://r0k.us/graphics/kodak/}}                    \\\hline
	\end{tabular}
	\label{tbl:dtinfo}
\end{table*}

\subsubsection{Zurich RAW to RGB (ZRR)}

\rodrigo{

ZRR dataset was proposed by Ignatov Et. al. ~\cite{Ignatov2019AIM2C} in order to get a large-scale real-world dataset, that deals with the task of converting original RAW photos captured by smartphone cameras to superior quality images achieved by a professional DSLR camera. The proposed database is publicly available and amounts to 22 GB, containing 20K real images captured synchronously by a Canon 5D Mark IV DSLR camera and Huawei P20 phone, in a variety of places and in various illumination and weather conditions. The photos were captured in automatic mode, however, some RAW–RGB image pairs are not perfectly aligned, which requires a preprocessing and matching performance. ZRR was a recurrent dataset choice for some RAW to RGB mapping problem works considered in this survey ~\cite{mei2019higherresolution, gupta2021delnet, ignatov2020replacing, zhang2021learning, dai2020awnet, kim2020pynetca}.

}

\subsubsection{Urban 100}

\rodrigo{

The Urban 100 dataset consists of 100 High-Resolution images with an assortment of real-world urban scenarios and structures. It was proposed by Huang Et. al. ~\cite{Jia2015Urban}, in order to address the lack of High-Resolution Datasets with indoor, urban, and architectural scenes. Urban 100 was constructed with synthetic images from Flickr\footnote{https://www.flickr.com/}, under Creative Commons license, resulting in a 1.14 GB dataset. It is a well-known public database for super-resolution tasks ~\cite{kim2020deep, guo2020residual, sun2020learning, a2021beyond, 9156993, kim2020deep}.

}

\subsubsection{McMaster dataset}

\jhessica{

The McMaster dataset was proposed by Zhang et al.~\cite{Zhang2011Color} and consists of eighteen sub-images of size 500x500 captured by Kodak film and then digitized. The sub-images were cropped from eight high-resolution natural images with size 2310x1814.  When compared with the Kodak color image dataset, McMaster shows images with more saturated colors and more color transitions between the image's objects. However, this dataset is still limited in scene variation and colors gradations. The McMaster dataset is used for color demosaicing in some of the articles in this survey~\cite{a2021beyond, guo2020residual, kim2020deep, DRDN, qian2021rethinking}.}

\subsubsection{Kodak}

\matheus{

Kodak \footnote{http://r0k.us/graphics/kodak/} is a little dataset composed of 24 photographic quality images of size 768x512 or 512x768 with a large variety of locations and lighting conditions. Initially, it was created to be a sample Kodak Photo CD, with some images captured by Kodak's professional photographers and others selected from the winners of the Kodak International Newspaper Snapshot Awards(KINSA). This dataset contains raw images in photo-cd (PCD) format and PNG format with 24 bits per pixel. Besides, many works use the Kodak dataset for compression tests and to validate methods that do tasks like demosaicking, denoising, and full ISP pipeline \cite{a2021beyond, guo2020residual, DRDN, kim2020deep, qian2021rethinking}.
}

\subsection{Papers Analysis}
\label{papers_analysis}

\matheus{The analysis of the works in some fields of study is fundamental to discover new ways to its evolution and improvement in future works. In this survey, the papers are analyzed about the following points:}
\begin{itemize}
    \item \matheus{\textbf{Details of method}:
The analysis of details of many methods can bring new ideas and identification of problems that future works can propose to solve.}

    \item \matheus{\textbf{Used datasets:}The camera hardware has many nuances and generates your own noise that is hard to simulate. Then the use of an appropriate dataset to train and validate the work is an important part to consider in creating a new ISP method. }

    \item \matheus{\textbf{Preocupation with computational cost:} The computational cost is an important point to consider in almost all applications of ISP, mainly used in embedded systems and mobile devices.}

    \item \matheus{ \textbf{Method evaluation}: How the method was evaluated may be well planned to indicate the contribution of this work in a determined study area. }
\end{itemize}

\section{Results}
\matheus{In this section, we analyzed the works on a quantitative comparison and indicated possible reasons for a method to present better than others. We divided this section into four subsections, where each one discussed results obtained in a dataset. In the first, we discussed the Zurich RAW to RGB dataset, which contains pairs of images from differents câmeras and misalignment between the pairs.  In the second, we discussed the Urban 100 dataset, which is composed of 100 High-Resolution images with real-world urban scenarios and structures. Then, in sequence, we discussed the works in the McMaster dataset, with 18 images captured by Kodak film. Finally, in the fourth section, we discussed the Kodak dataset with 24 photographic quality images of size 768x512 or 512x768, generally used in compression tests and to validate methods for tasks like demosaicking and denoising, etc.}

\begin{minipage}[t]{0.5\textwidth}
\centering
	\begin{threeparttable}
	\centering
	\caption{ZURICH RAW2RGB dataset~\cite{ignatov2020replacing}}
	\label{tbl:zurich_dataset}
	\vspace{0.5cm}
	\begin{tabular}{l|c|c}\hline
		\textbf{Networks}                    & \textbf{PSNR} & \textbf{SSIM}\\\hline%
		Del-Net\cite{gupta2021delnet}        & 21.46         & 0.745         \\
		PyNet\cite{ignatov2020replacing}     & 21.19         & 0.746         \\
		AWNet (Ensemble)\cite{dai2020awnet}             & 21.86         & 0.781        \\
		AWNet (Demosaiced)\cite{dai2020awnet}             & 21.38         & 0.745        \\
		AWNet (RAW)\cite{dai2020awnet}             & 21.58         & 0.749        \\
		PyNet-CA\cite{kim2020pynetca}        & 21.50         & 0.743         \\
		LiteISPNet\cite{zhang2021learning}   & 21.55         & 0.748         \\
\hline
		LiteISPNet\cite{zhang2021learning}   & \textbf{23.76}\tnote{a}         & \textbf{0.873}\tnote{a}      \\  %
		HERNet\cite{mei2019higherresolution} & 22.59\tnote{b}         & 0.81\tnote{b}       \\  %
		\hline
		\end{tabular}
		\begin{tablenotes}
			\footnotesize
			\item[a] Align ground truth with RAW image.
			\item[b] Data with diferent distribution.

		\end{tablenotes}
	\end{threeparttable}

\end{minipage}
\begin{minipage}[t]{0.5\textwidth}
\centering

	\begin{threeparttable}
		\centering
		\caption{Urban 100 dataset~\cite{Jia2015Urban}}
		\label{tbl:urban100}
		\begin{tabular}{l|c|c}\hline
			\textbf{Networks}                             & \textbf{PSNR}  & \textbf{SSIM}  \\\hline%
			DPN~\cite{kim2020deep}                    & 37.70  & 0.9799\\
			\hline
			PIPNet~\cite{a2021beyond}                     & 37.51\tnote{a} & 0.9731\tnote{a}\\ %
			TENet~\cite{qian2021rethinking}                & 29.37\tnote{c} & 0.9061\tnote{c}\\
			SGNet\cite{9156993}                           & 34.54\tnote{a} & 0.9533\tnote{a}\\ %
			RLDD~\cite{guo2020residual}      & \textbf{39.52}\tnote{b}          & \textbf{0.9864}\tnote{b} \\ %
			\multicolumn{1}{p{3cm}|}{\raggedright RestoreNet w/ PatchNet~\cite{sun2020learning}} & 34.66\tnote{a} &        -       \\  %
			DPN~\cite{kim2020deep}                    & 37.70  & 0.9799\\

			\hline
		\end{tabular}
		\begin{tablenotes}
			\footnotesize
			\item[a] Data with noise.
			\item[b] 10 pixels were removed from image's border to calculate the PSNR value.
			\item[c] Downsampled data to do demosaicing + SR task.
		\end{tablenotes}
	\end{threeparttable}

\end{minipage}

\subsection{Zurich RAW to RGB}
\matheus{The HERNet and LiteISPNet present versions with better results in the \autoref{tbl:zurich_dataset}, but it uses, respectively, a different distribution of data and the RAW image aligned with the ground-truth. With the remain of results, the AWNet outperformed other methods in PSNR and SSIM scores, and the use of a self-ensemble strategy can explain this, as RAW and Demosaiced versions have scores relatively much lower than the ensembled version. Besides, the AWNet RAW version has slightly higher results than LiteISPNet and PyNet-CA, which can result from the wavelet transform and context global blocks.  Furthermore, the LiteISPNet has impressive results with introducing an additional method to align the images, improving network training.}

\subsection{Urban 100}
\matheus{Urban 100 dataset is used for many tasks of image restoration. Then the works do modifications to adequate this dataset to a specific task. As shown in Table~\ref{tbl:urban100}, the  RLDD had the highest scores compared to other methods but was validated with the image's border removed, and this factor can help increase scores. Furthermore, the PIPNet, even being validated with noisy data, has comparable results with DPN. The introduction of attention mechanisms that bring good correlations in relation to depth and spatial dimensions may be the cause of these promising results in this dataset.}

\subsection{McMaster}
The DRDN+ and DRDN methods showed better results in the PSNR metric in Table~\ref{tbl:mcmaster_dataset}. These methods are CNN-based models and focus on demosaicking. The DPN method, which also focused on demosaicing, showed better results in the SSIM metric and better artifacts reduction in its qualitative evaluation. The TENet network had the lowest performance in the PSNR and SSIM metrics, however, it is worth noting that the goal of this network is to make an entire ISP pipeline enhancement, instead of only the demosaicing task.

\begin{minipage}[t]{0.5\textwidth}
\centering

	\begin{threeparttable}
		\centering
		\caption{McMaster dataset~\cite{Zhang2011Color}}
		\label{tbl:mcmaster_dataset}
		\begin{tabular}{l|c|c}\hline
			\textbf{Networks}                        & \textbf{PSNR}  & \textbf{SSIM} \\\hline  %

			DRDN~\cite{DRDN}                         & 38.88          & 0.9689          \\
			DRDN+~\cite{DRDN}                        & \textbf{39.02}          & 0.9697          \\
			DPN~\cite{kim2020deep}                   & 37.6           & \textbf{0.9842}          \\
			\hline
			TENet\cite{qian2021rethinking}           & 32.40\tnote{c}          & 0.9163\tnote{c}          \\
			PIPNet~\cite{a2021beyond}                & 38.13\tnote{a} & 0.9612\tnote{a}\\ %
			RLDD~\cite{guo2020residual} & 36.61\tnote{b}          & 0.9725\tnote{b} \\        %
			\hline
		\end{tabular}
		\begin{tablenotes}
			\footnotesize
			\item[a] Data with noise.
			\item[b] 10 pixels were removed from image's border to calculate the PSNR value.
			\item[c] Downsampled data to do demosaicing + SR task.
		\end{tablenotes}
	\end{threeparttable}
\end{minipage}
\begin{minipage}[t]{0.5\textwidth}
\centering

	\begin{threeparttable}
		\centering
		\caption{Kodak Dataset~\cite{kodakDataset}}
		\label{tbl:kodak}
		\begin{tabular}{l|c|c}\hline
			\textbf{Networks}                        & \textbf{PSNR} & \textbf{SSIM}    \\\hline%

			DRDN~\cite{DRDN}                         & 42.43          & 0.9889         \\ %
			DRDN+~\cite{DRDN}                        & 42.66          & \textbf{0.9893}         \\ %
			DPN~\cite{kim2020deep}                   & 40.1           & 0.9846         \\ %
			\hline
			TENet\cite{qian2021rethinking} & 31.39 \tnote{a} & 0.8965\tnote{a}  \\
			PIPNet~\cite{a2021beyond}                & 39.37\tnote{a} & 0.9768\tnote{a} \\
			RLDD~\cite{guo2020residual} & \textbf{42.76}\tnote{b}          & 0.9893 \\
			\hline
		\end{tabular}
		\begin{tablenotes}
			\footnotesize
			\item[a] Data with noise.
			\item[b] 10 pixels were removed from image's border to calculate the PSNR value.
		\end{tablenotes}
	\end{threeparttable}
\end{minipage}

\subsection{Kodak}

\rodrigo{Table~\ref{tbl:kodak} shows the reviewed paper results in the Kodak dataset \footnote{http://r0k.us/graphics/kodak/}. RLDD~\cite{guo2020residual} achieved the best PSNR metric performance, whereas DRDN+~\cite{DRDN} had the best SSIM metric performance. The RLDD framework combines Denoising and Demosaicing techniques, delivering proper quantitative and qualitative results. It is important to reinforce that RLDD authors removed 10 pixels from the Kodak image's borders to calculate the PSNR. TENet and PIPNet introduced artificial noise models into the dataset for deeper denoising study. DRDN stands out in terms of efficiency and accuracy, in large part because of its block-wise convolutional neural networks which consider local features and a sub-pixel interpolation layer.}

\subsection{Source Code Links}

\matheus{The Table~\ref{t.sourcecode} presents the links to the source codes of some works mentioned in this survey.}

\begin{table*}[htb!]
\small
\centering
\caption{Summarization of some approaches considered in the survey and their respective source codes.}
\begin{tabular}{c|ll}
\textbf{Short Name }     & \multicolumn{1}{c}{\textbf{Source Code}} \\\hline
 CycleISP \cite{Zamir2020CycleISP}                             &PyTorch: \url{https://github.com/swz30/CycleISP} \\\hline
CURL \cite{moran2021curl}                             & PyTorch: \url{https://github.com/sjmoran/CURL}\\\hline
 LiteISPNet \cite{zhang2021learning}                             &PyTorch: \url{https://github.com/cszhilu1998/RAW-to-sRGB}\\\hline
PyNet \cite{ignatov2020replacing}                             &  \begin{tabular}[c]{@{}l@{}}PyTorch and Tensorflow: http://people.ee.ethz.ch/~ihnatova/pynet.html \end{tabular}  \\\hline
 PyNet-CA \cite{kim2020pynetca}                             & PyTorch: \url{https://github.com/egyptdj/skyb-aim2020-public} \\\hline
BJDD \cite{a2021beyond}                              & PyTorch: \url{https://github.com/sharif-apu/BJDD_CVPR21}\\\hline
TENet \cite{qian2021rethinking}                        &  PyTorch: \url{https://github.com/guochengqian/TENet}  \\\hline
 AWNet \cite{dai2020awnet}                        & \begin{tabular}[c]{@{}l@{}}PyTorch: https://github.com/Charlie0215/AWNet-Attentive\\-Wavelet-Network-for-Image-ISP \end{tabular}  \\\hline
Bayer Methods \cite{bayermethods}                        &PyTorch: \url{https://github.com/Jiaming-Liu/BayerUnifyAug}  \\\hline
 HERNet \cite{mei2019higherresolution}                        & PyTorch: \url{https://github.com/MKFMIKU/RAW2RGBNet}  \\\hline
\end{tabular}

\label{t.sourcecode}
\end{table*}

\section{Discussion}
\label{s.discussion}

\matheus{We provide, in this section, points of improvement in these works about its methodology and evaluation. However, much of it has highlights which can contribute a lot about the learned cameras ISPs. This section will address the negative and positive points of the works described above and ask questions that may evolve into new paths to rods in this research field.}

\subsection{Dataset Created with Different Cameras}

\matheus{Some datasets ~\cite{ignatov2020replacing, ignatov2021learned} have the pair images captured by different cameras, where one of the main objectives of the works that use these datasets is to learn the features of high-quality cameras and apply them in cameras with more limitations. Because of the differences between the devices, these datasets have alignment problems about the image patch pairs. Doing the RAW to RGB mapping can be challenging because the patch pixels do not correspond to theirs pixels. The trivial solution is to align the pairs, but how to do this is the main question. Zhang et al.\cite{zhang2021learning} propose the use of Deep Learning where outperforms the models trained with the misalignment dataset. Another way that many works \cite{ignatov2021learned, dai2020awnet, ignatov2020replacing, gupta2021delnet} use to mitigate this problem is to adapt the loss function to not punish the models where the generated image is misaligned with the ground truth. Models trained with losses that try to get closer to human perception are more invariant to misalignment and generate images with more details. Currently, the metrics that are closest to this make use of Deep Learning techniques, using extracted features of images with the use of pre-trained Deep Learning models \cite{zhang2018unreasonable, mustafa2021training}. Thus comes the doubt of how the best misaligment solution should be used to mitigate the problem adding no overhead on the computational cost of the training process. This problem can become a search point by itself, such as alignment methods to RAW and RGB images of differents cameras, facilitating the creation of new datasets and improving the performance of the new works.}

\subsection{Methodologies}
\matheus{Seen as a point that helps share knowledge and the research field evolution is the justification of each stage, from architecture construction to validation of the work or the choice of datasets. Nevertheless, some works do not do it. For example, Liu et al. \cite{9156993} do not explain well why the use of some datasets and Liu et al.~\cite{bayermethods} do not indicate the motivation to use the U-Net model. Besides, many methods do not make the source code available in its paper \cite{mei2019higherresolution, DRDN, liang2019cameranet, ratnasingam2019deep, deepedge, guo2020residual}, making it difficult for future validations and comparisons with other methods and making the validation less reliable.}

\subsection{Evaluation Protocols}

\matheus{Train models using synthesized RAW image data when RAW and RGB images are scarce in the dataset is a good way to increase generalization. However, this cannot be a great option to validate methods proposed to map ISP because the synthesized RAW image data results from interpolation of RGB processed images, has pixels with less information, and the resulting noises of camera hardware are complex and hard to generate. Nevertheless, many works use datasets created for tasks like denoising, deblurring, or super-resolution with the generation of RAW images from these datasets that disbelieve the applicability of these methods in the real world. This problem raises the question of how to create RAW to RGB datasets with good representation and high quality for further research. }

\matheus{On the other hand, few works tested the models in embedded systems or mobile devices. This can be a problem because most applications of ISPs are in devices with low computational capacity, and the fault of validation in this way can not prove the possible use in real applications. In view of this, the Mobile 2021 Challenge released the task: "Learned Smartphone ISP on Mobile NPUs with Deep Learning."\cite{ignatov2021learned} It considers the processing and time of execution in mobile NPUs and the image quality like the other two previous challenges. This proposal can encourage future works to care more about these points.}

\section{Conclusion}
\label{s.conclusion}

The ISP pipeline is an important combination of techniques, essential for the creation of quality digital images from camera sensors.
This survey provided an in-depth research on the applications of deep learning techniques to ISP tasks,
regarding the application of networks for solving partial steps or the complete pipeline.
Additionally, it also provided an introduction to both ISP and deep learning areas, alongside with a
detailed overview of software ISP, regarding its fundamentals and individual steps.

The works surveyed in this paper were selected based on their novelty, their target task, and the applied deep learning techniques.
Among the 27 reviewed papers, $30\%$ had applied DNNs for replacing the complete ISP pipeline. This reveals a new trend that explores the generalization
ability of CNNs to learn all the ISP individual tasks, aiming to apply all of them in a single forward operation.

Furthermore, this work also summarized five datasets often used for ISP tasks: Urban 100~\cite{Jia2015Urban}, ZRR dataset~\cite{Ignatov2019AIM2C}, Kodak\footnote{http://r0k.us/graphics/kodak/},
 and McMaster dataset~\cite{Zhang2011Color}. The availability of quality datasets is necessary for the research and development of
new solutions to any deep learning application area. In this scenario, new ISP-related datasets can improve some limitations existent in the surveyed datasets.

Finally, during the development of this survey, some critical points were detected and are highlighted below:

\begin{itemize}
    \item The use of different cameras for producing RAW-to-RGB datasets creates alignment issues that require the application of additional
    techniques during the training of DNNs. These mitigators can add computational costs and interference in the overall performance of the networks.
    \item Better ablation studies, regarding the effect of each architectural component present in proposed solutions, alongside with the distribution of the source code,
    can facilitate the future development of new techniques and the overall improvement of the research area.
    \item As more approaches are proposed to the ISP pipeline replacement task, a more consistent evaluation procedure, alongside with the definition of common target datasets,
    will facilitate the comparison between methods, helping to define state-of-the-art performance.
    \item Besides being one of the main target applications, mobile application for networks that replace the complete ISP pipeline are not often discussed. A more in-depth
    evaluation of the proposed methods performance on edge devices is important to identify components that can be optimized or new techniques focused on these target environments.
\end{itemize}

As future work, we intend to pursue two main approaches to Deep Learning-based ISP applications:

\begin{itemize}
    \item The development of a complete ISP pipeline network focused on execution in a Raspberry Pi board, aiming to explore the deployment on edge challenge.
    \item The application of vision transformers to the complete ISP task, aiming to explore the good results achieved by transformers in other computer vision tasks.
\end{itemize}

\bibliographystyle{unsrt}  
\bibliography{survey-isp-arxiv}

\end{document}